\documentclass[sigconf]{acmart}

\def\BibTeX{{\rm B\kern-.05em{\sc i\kern-.025em b}\kern-.08emT\kern-.1667em\lower.7ex\hbox{E}\kern-.125emX}}
\pdfoutput=1
    
\copyrightyear{2019} 
\acmYear{2019} 
\setcopyright{acmlicensed}
 \acmConference[NICE '19]{Neuro-inspired Computational Elements Workshop}{March 26--28, 2019}{Albany, NY, USA}
\acmBooktitle{Neuro-inspired Computational Elements Workshop (NICE '19), March 26--28, 2019, Albany, NY, USA}
\acmPrice{15.00}
\acmDOI{10.1145/3320288.3320293}
\acmISBN{978-1-4503-6123-1/19/03}

\begin{document}

%
\title{Signal Conditioning for Learning in the Wild}

%

\author{Ayon Borthakur}
\affiliation{
  \institution{Field of Computational Biology, Cornell University}
  \city{Ithaca}
  \state{NY}
  \postcode{14853}
  \country{USA}}
\email{ab2535@cornell.edu}

\author{Thomas A. Cleland}
\affiliation{
  \institution{Dept. Psychology, Cornell University}
  \city{Ithaca}
  \state{NY}
  \postcode{14853}
  \country{USA}}
\email{tac29@cornell.edu}
\orcid{0000-0001-7506-1201}

%
\renewcommand{\shortauthors}{Borthakur and Cleland}

%
\begin{abstract}
The mammalian olfactory system learns rapidly from very few examples, presented in unpredictable online sequences, and then recognizes these learned odors under conditions of substantial interference without exhibiting catastrophic forgetting. We have developed a brain-mimetic algorithm that replicates these properties, provided that sensory inputs adhere to a common statistical structure.  However, in natural, unregulated environments, this constraint cannot be assured. We here present a series of signal conditioning steps, inspired by the mammalian olfactory system, that transform diverse sensory inputs into a regularized statistical structure to which the learning network can be tuned. This preprocessing enables a single instantiated network to be applied to widely diverse classification tasks and datasets - here including gas sensor data, remote sensing from spectral characteristics, and multi-label hierarchical identification of wild species - without adjusting network hyperparameters. 
\end{abstract}

%
%

%

%

%
\maketitle
\section{Introduction}
\begin{figure}[h]
  \centering
  \includegraphics[width=\linewidth]{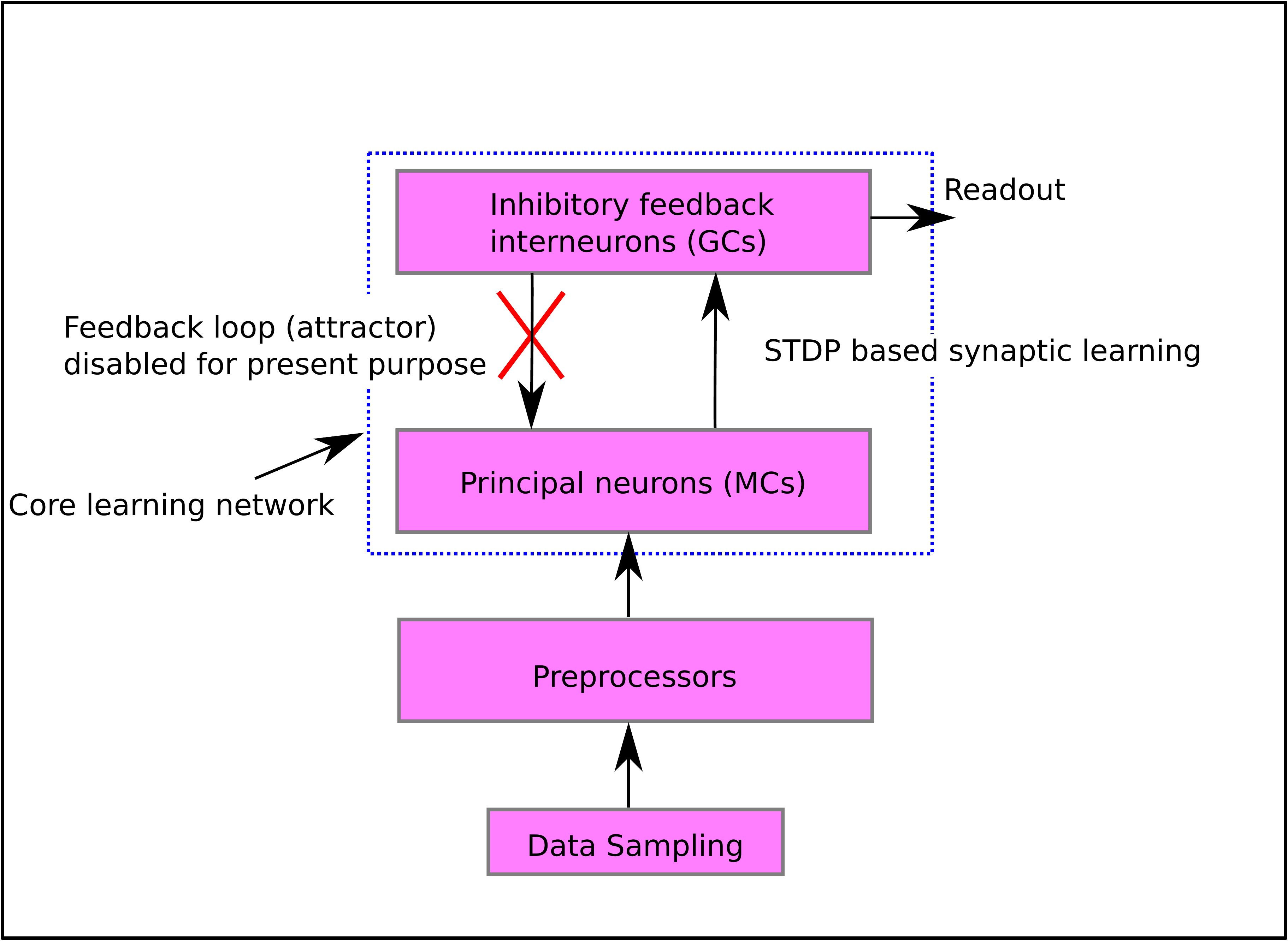}
  \caption{Schematic overview of brain-mimetic model. Preprocessors include multiple signal conditioning functions attributed to glomerular layer circuitry in the biological system, including normalization, contrast enhancement, and statistical regularization.  The core learning network comprises an inhibitory feedback loop between principal neurons and interneurons in which sensory information is conveyed by the phases of principal neuron spike times with respect to the underlying gamma cycle; learned patterns form attractors that classify test samples \cite{imam_rapid_2019}.  For present purposes, this inhibitory feedback was disabled and the patterns of interneuron activation were read out directly.  Classification was estimated based on the minimum Hamming distance between test sample and learned ensembles in the interneuron representation.}
  \label{schematic}
\end{figure}

The mammalian olfactory system learns and recognizes odors of interest under suboptimal circumstances and in unpredictable environments.  Real-world odor stimuli vary in their concentrations and qualities, and are typically encountered in the presence of unpredictable configurations of competing background odors that can substantially occlude the profile of sensory receptor activation on which odor quality recognition nominally depends.  Moreover, odor learning is rapid, and multiple odors can be learned in arbitrary sequences (online learning) without their learned representations interfering with one another (causing catastrophic forgetting) and without training data being somehow stored to maintain or restore learning performance.  Altogether, this suite of sensory sampling challenges constitutes the problem of \textit{learning in the wild}. 

We have designed and implemented a spiking neural network (SNN) algorithm for learning and identifying chemosensor array responses and other intrinsically higher-dimensional signals, based on the architecture of the mammalian main olfactory bulb (MOB) \cite{imam_rapid_2019, borthakur_spike_2019}.  Briefly, primary chemosensory neurons expressing a single type of receptor converge to common locations on the MOB surface, there forming clusters of neuropil called glomeruli.  Activity in these glomerular networks then is sampled and processed by second-order principal neurons and multiple classes of interneurons. Glomerular activation profiles across hundreds of receptor types (\textasciitilde1200 in rodents) constitute high dimensional vectors describing odor qualities embedded in multiple sources of noise.  

Importantly, glomerular-layer network interactions perform multiple signal conditioning tasks on raw chemosensory inputs.  Recognizing odor stimuli across wide concentration ranges, for example, depends on the coordination of multiple computational elements \cite{cleland_construction_2014, cleland_sequential_2011}, including a global inhibitory feedback loop within the MOB glomerular layer that limits concentration-dependent heterogeneity in the activity of MOB principal neurons \cite{cleland_non-topographical_2006, cleland_relational_2007, banerjee_interglomerular_2015}.  A version of this input normalization algorithm has been implemented previously on the IBM TrueNorth neuromorphic hardware platform \cite{imam_implementation_2012}. 

Our present SNN algorithm for machine olfaction, implemented on Intel Loihi, learns rapidly from one or few shots, resists catastrophic forgetting, and classifies learned odors under high levels of impulse noise \cite{imam_rapid_2019}.  Moreover, the interpretability of the algorithm enables the causes of the classification to be ascertained post hoc, in principle enabling the identification of the specific combinations of input features that determine a sample's classification. Subsequent generalized versions of this model under development relax control over key parameters in order to develop an experience-dependent metric of similarity for purposes of hierarchical classification.  However, the plastic network at the core of this generalized algorithm is sensitive to the statistical parameters of sensory input, potentially requiring parameter retuning in order to maintain effective classification performance when the input statistics change.  We sought instead to implement a consistent set of adaptive signal conditioning mechanisms that would enable any sensor array input profile to be accepted by a given instantiated network for learning and high-fidelity classification under noise without requiring parameter retuning.  This strategy enables multiple, statistically diverse input signals to each be encountered, learned, and classified by the same network - an essential capacity for an artificial sensory system deployed into an unknown \textit{wild} environment.

\section{Algorithm adaptation for learning in the wild}
The feedback loop comprising the core network recruits populations of interneurons during learning to represent higher-order stimulus features \cite{imam_rapid_2019}.  To explicitly represent stimulus similarity (a prerequisite for constructing hierarchical representations on this metric), these recruited populations must be permitted to overlap in their representation of similar input stimuli - a goal that requires relaxing control over interneuron recruitment. However, this poses a challenge, in that differently structured sensory inputs can be poorly suited for the parameterized network.  Sensors in the array that are mismatched to the environment or to one another, sensory input profiles that differ substantially in mean amplitudes (e.g., higher or lower analyte concentrations), or even input profiles that are broader and flatter or steeper and narrower than expected - all have the potential to disrupt learning and classification performance.  In lieu of retuning network hyperparameters, we sought to construct a network architecture that could learn and classify input patterns irrespective of their statistical properties.  That is, \textit{learning in the wild} requires that a single parameterized network be able to learn and classify any set of relevant signal patterns that it may encounter. 

We here present two elements of network architecture, inspired by the biological olfactory system, that enable learning in the wild.  First, we present a series of signal conditioning \textbf{preprocessors}, based on elements of MOB glomerular-layer circuitry, that effectively normalize and regularize sensory input patterns.  Second, we show that the implementation of \textbf{heterogeneity} in key network parameters further broadens network tolerance and improves classification performance.  To illustrate these effects more clearly, we omit the inhibitory feedback loop that governs the attractor dynamics of the core learning network \cite{imam_rapid_2019}, and instead report an intermediate estimate of classification accuracy derived from the first projection of the preprocessed input stream onto the interneurons of the core network (i.e., the \textit{EPLff} component described in \cite{borthakur_spike_2019}; Figure~\ref{schematic}).  We also report the profiles of interneuron recruitment as an indicator of the statistical similarities among input signals after preprocessing, and by extension the adaptiveness of these representations for the fixed hyperparameters of the core learning network. 

\subsection{Preprocessors for signal conditioning}
We implemented three preprocessors that were applied in sequence to sampled input vectors. Among these, the second (intensity normalization) is directly inspired by glomerular-layer operations in the MOB \cite{cleland_non-topographical_2006, cleland_relational_2007, cleland_construction_2014, banerjee_interglomerular_2015, whitesell_interglomerular_2013}, and the third (heterogeneous duplication) makes use of known circuit motifs in the MOB \cite{gire_etcells_2012} to which no clear function has previously been attributed.  
\subsubsection{Sensor scaling}
Sensor scaling enables the inclusion of heterogeneous sets of sensors or feature values that may be drawn from different scales of measurement.  Based on a small sample of inputs (validation set), this preprocessor estimates the range of values received from each sensor and scales each sensor value accordingly.  Because samples cannot be guaranteed to include the full range of values that a sensor may deliver, this step does not comprise idealized scaling, but order-of-magnitude approximate scaling that prevents a subset of inputs from inappropriately dominating network plasticity.  To enhance feature value differences, the scaled parameters then are multiplied by an equidimensional vector with values drawn from a uniform distribution between $0.5$ and $1.0$; once defined, these vector values are a constant attribute of an instantiated network.   

\subsubsection{Unsupervised intensity normalization}
For some input streams, stimulus intensity can interfere with identity.  For example, increased concentrations of chemical analytes will nonuniformly increase the responses of array chemosensors, which impairs analyte recognition across concentrations.  In the biological system, it has been proposed that multiple coordinated mechanisms serve to reduce the impact of intensity differences (i.e., yielding \textit{concentration tolerance}, or \textit{concentration invariance}), with the remaining uncompensated intensity effects being learned as part of the characteristic variance of that stimulus \cite{cleland_sequential_2011}.  We adopted this principle, implementing a nonspecific inhibitory feedback mechanism inspired by the deep glomerular layer of the olfactory system \cite{cleland_relational_2007, cleland_construction_2014, banerjee_interglomerular_2015} and comparable to one previously implemented in neuromorphic hardware \cite{imam_implementation_2012}.  This preprocessor enables the recognition of odorant signatures presented at a range of untrained concentrations, even under few-shot learning conditions \cite{borthakur_spike_2019}.  Intensity normalization in the biological system also is required for regulated high-dimensional contrast enhancement \cite{cleland_non-topographical_2006, cleland_construction_2014, cleland_early_2010}, although the latter algorithm was not incorporated into the present simulations.  
\subsubsection{Heterogeneous duplication}
\begin{figure}[h]
  \centering
  \includegraphics[width=\linewidth]{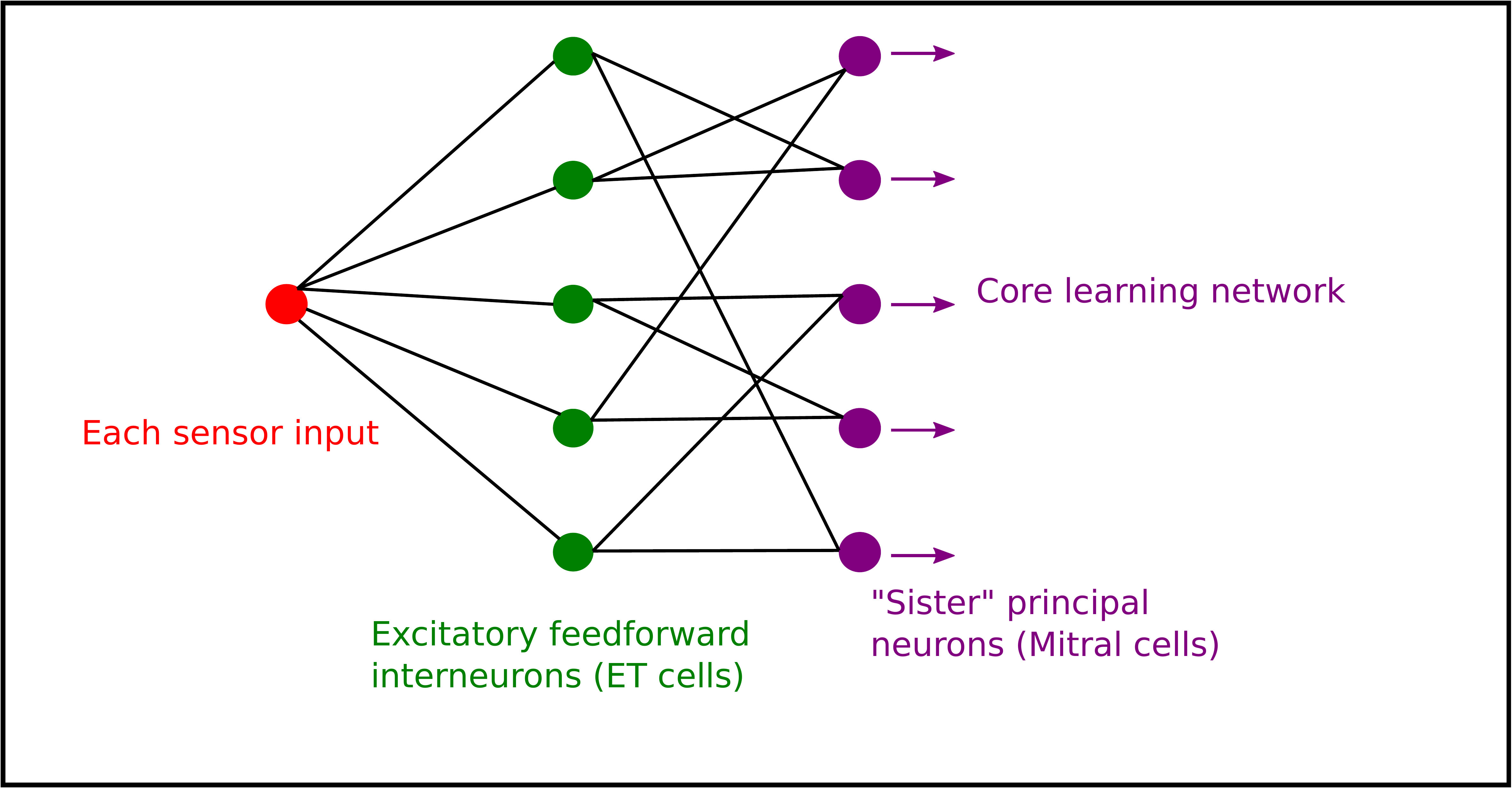}
  \caption{Heterogeneous duplication preprocessor network. The brain-mimetic implementation of heterogeneous duplication is modeled after aspects of the intraglomerular circuitry of the mammalian olfactory bulb \cite{gire_etcells_2012}, and serves to statistically regularize the distribution of amplitudes among inputs.  Each sensor input is delivered to a number of excitatory feedforward interneurons (here, five) comparable to the external tufted (ET) cells of the OB, and from there, via sparse, random, feedforward projections, to the principal neurons of the core learning network (analogous to OB mitral cells). This preprocessor expands the size of the core learning network; here, each sensor now corresponds to  a column with five computing units/sister mitral cells (MCs).}
  \label{het_dup}
\end{figure}

Despite sensor scaling and intensity normalization, the different distributions of activity levels across the array of inputs still could disrupt the performance of the core attractor  in the generalized network model, most prominently by recruiting widely divergent numbers of interneurons during learning.  To address this problem without resorting to retraining network hyperparameters, we duplicated each input across a number of excitatory feedforward interneurons (e.g., five) and then randomly projected the activity of these interneurons onto a similar number of principal neurons (Figure~\ref{het_dup}).  Because the number of processing columns of the core learning network is determined by the number of principal neurons, this also expanded the dimensionality of the network. The integration and synaptic properties of both cell types were heterogeneous across the duplicates, drawn randomly from a defined range during network instantiation.  This feedforward heterogeneous duplication with random projections regularized the statistical distribution of input levels into a consistent range (details in \textit{Empirical Results}), enabling a single parameterization of the core network to be effective across a wide range of poorly-behaved inputs. 

Interestingly, the need for statistical regularization of afferent input activity has not yet been recognized as a problem in the biological olfactory system.  It may be that the biological system is tolerant of statistically diverse inputs via other mechanisms that have yet to be elucidated, but it is nevertheless intriguing that this feedforward projection motif is the dominant mechanism of sensory sampling in the biological olfactory bulb.  Specifically, convergent primary sensory neurons primarily excite external tufted (ET) cells within a glomerulus (along with inhibitory periglomerular cells), and these ET cells then in turn excite the principal neurons of that glomerulus.  This indirect pathway has been shown to be the dominant path of afferent excitation, with direct OSN-to-principal neuron excitation being relegated to a considerably smaller role \cite{whitesell_interglomerular_2013}. 

\subsubsection{Goodness of preprocessing metric, $g_{p}$}
The preprocessor sequence described above regularized widely diverse input signals into a common statistical distribution to which the core network was optimized.  Well-regularized sensory inputs recruit consistent numbers of interneurons into the representation during learning, and activate appropriate interneuron ensembles during testing. To assess the functional adequacy of preprocessing, we developed a \textit{goodness of preprocessing} metric, $g_{p}$, as a measure of the consistency of interneuron recruitment efficacy across a heterogeneous range of samples: 
\begin{equation}
  g_{p} = min(min(v), 1) * \frac{\sum{\frac{v_{i}}{max(v)}}}{\dim{v}}
  \label{gp}
\end{equation}

\noindent where $v$ is an integer vector of interneuron spike counts and dim $v$ denotes the number of samples under consideration.  This equation has two factors.  First, the \textit{no-spike penalty}
\begin{equation}
    min(min(v), 1)
\end{equation}
is zero if any of the stimuli presented fail to activate any interneurons at all; otherwise its value is unity.  Second, the \textit{interneuron activation similarity index} 
\begin{equation}
     \frac{\sum{\frac{v_{i}}{max(v)}}}{\dim{v}}
\end{equation}
reflects the similarity of interneuron recruitment levels across all stimulus presentations (i.e., across multiple different stimuli, potentially also including a range of stimulus intensities or concentrations). These two factors together generate a value of $g_{p}$ between 0 and 1. A $g_{p}$ value approaching unity indicates that all test stimuli activate approximately the same nonzero number of interneurons; lower values indicate that different stimuli recruit substantially different numbers of interneurons, or none at all, which may impair the performance of a given core network for some of these stimuli.

\subsection{The core learning network}
The core learning network comprises a recurrent excitatory-inhibitory feedback loop between populations of principal neurons (\textit{mitral cells}, MCs) and inhibitory interneurons (\textit{granule cells}, GCs) (Figure ~\ref{schematic}). Rapid online learning progressively modifies the synaptic weights of this network, generating attractors that correctly classify even highly degraded, noisy inputs \cite{imam_rapid_2019}. Importantly, this feedback loop recruits interneurons during learning; to model similarity, a prerequisite for constructing hierarchical representations, these recruited populations must be permitted to overlap in their representation of similar input stimuli. This renders the network more sensitive to the statistics of sensory input, thereby requiring signal conditioning if parameter retuning is to be avoided.  As noted above and in Figure ~\ref{schematic}, to focus on the statistical regularization of sensory inputs to this learning network, we include only the feedforward portion of the core network in the present simulations, reading out  regularization and intermediate classification results (i.e., the goodness of preprocessing index $g_{p}$ and thresholded Hamming distances) directly from the interneuron population.  As in the intact model \cite{imam_rapid_2019}, sensor activation levels are represented in principal neurons by a spike phase code \cite{panzeri_sensory_2010, linster_decorrelation_2010} with respect to an underlying gamma oscillation, and we use an asymmetric spike timing-dependent plasticity (hSTDP) learning rule to modify MC-to-GC synaptic weights.

\subsubsection{Heterogeneity in model parameters}
\- Heterogeneity is abundant in biology; information is commonly represented in populations of neurons with similar but not identical properties. This is often elided as unavoidable biological variability, but may in fact serve an important computational purpose.  For example, recent experimental studies in the retina have shown that the population code exhibited by a heterogeneous ensemble of neurons is considerably more reliable than that of a homogeneous ensemble \cite{berry_ii_functional_2018}. To assess and take advantage of this potential, we here incorporate network heterogeneities in three ways: 
\begin{enumerate}
\item \textbf{Nonuniform sensor scaling:} This process is part of the sensor scaling preprocessor described above, employed to ensure feature value differences among inputs. 
\item \textbf{Heterogeneous duplication:} The heterogeneous duplication preprocessor fans out a common input stream to a heterogeneous population of $m$ excitatory feedforward interneurons, which then deliver this input to $n$ sister MCs via sparse random projections (Figure \ref{het_dup}).
\item \textbf{Model parameter heterogeneities:} We assigned variable spiking thresholds to sister MCs and to GC interneurons.  These partially redundant MC groups further enabled us to assign a wide range of MC-to-GC synaptic connection densities across the core learning network.  Finally, the maximum permitted synaptic weights $w_{max}$ under the STDP rule were heterogeneous; we refer to this overall rule as $hSTDP$.  These heterogeneities render the post-signal conditioning learning network more robust to statistically diverse datasets.  
\end{enumerate} 

\subsubsection{Heterogeneous spike timing-dependent plasticity (hSTDP)}
\- Per this learning rule, MC-to-GC excitatory synaptic weights were potentiated when MC spikes preceded GC spikes; otherwise these synapses were depressed. The $hSTDP$ rule parameters $a_{p}$, $a_{m}$, $tau_{p}$, $tau_{m}$, and $w_{scale}$ were tuned using a synthetic dataset \cite{borthakur_neuromorphic_2017, borthakur_spike_2019}, whereas the distribution of maximum synaptic weights $w_{max}$ was tuned only once using a validation set from Batch 1 of the UCSD chemosensor drift dataset \cite{vergara_chemical_2012, rodriguez-lujan_calibration_2014}. Training and testing with the additional datasets described herein also used this same instantiated, parameterized network.  

\section{Experimental datasets}
The results presented here were generated using a common network with all hyperparameters predetermined except for the number of columns and, in one case, the number of GC interneurons per sensor.  The number of processing columns depended directly on the number of sensor inputs provided by the dataset (input data dimensionality) multiplied by the divergence ratio $n$ of the heterogeneous duplication preprocessor (held constant at 5 for all simulations herein; Figure \ref{het_dup}).  Excitatory synaptic weights in the core network were plastic, governed by an $hSTDP$ rule with fixed parameters as described above.
\subsection{UCSD gas sensor drift dataset}
We first applied our algorithm to the publicly available UCSD gas sensor drift dataset \cite{vergara_chemical_2012, rodriguez-lujan_calibration_2014, noauthor_uci_nodate-2, ucirepo}, modestly reconfigured to assess online learning. The dataset contains 13910 measurements in total, taken from an array of 16 MOS chemosensors exposed to 6 gas-phase odorants presented across a substantial range of concentrations ($10-1000 ppmv$).  Most importantly, these data were gathered in ten batches over the course of three years; owing to sensor drift, the chemosensors' responses to odorants changed drastically over this timescale, presenting a challenge to classification algorithms that must model or otherwise compensate for that drift. For this study, we used data from Batches 1 (sensor age 1-2 months) and 7 (sensor age 21 months). As in previous work, we used only the peak sensor responses (16 out of the available 128 features in the dataset) for training and testing \cite{borthakur_spike_2019}. To better assess online learning, we reconfigured the dataset into six \textit{groups} corresponding to the six gas types, and trained the network with data from each of these six groups separately, in order.  Consequently, each training set comprised 1-10 samples (for 1-shot through 10-shot learning, respectively) of the same odorant, at randomly selected concentrations. After training on each odorant group, we tested all six odorants (at randomly selected concentrations) before proceeding to train the next group in the list, until the network had learned all six odorants.  Testing an odorant on which the network had not yet been trained produced the classification result \textit{none of the above} - a critical capability for \textit{learning in the wild}, wherein many presented odorants would be unfamiliar and should not be forced incorrectly into existing classes.  For sensor scaling and parameter tuning for this and all subsequent data sets, we used $10\%$ of the Batch 1 data as a validation set.  The six odorant groups, in the order of training, included ammonia (\textit{group 1}), acetaldehyde (\textit{group 2}), acetone (\textit{group 3}), ethylene (\textit{group 4}), ethanol (\textit{group 5}), and toluene (\textit{group 6}).  

\subsection{Forest type spectral mapping dataset}
This dataset is designed to identify forest types in Japan using spectral data from ASTER satellite imagery \cite{johnson_using_2012, noauthor_uci_nodate-1, ucirepo}. Each of the 326 samples includes 27 spectral features. We used $10\%$ of the data as a validation set for preprocessor scaling.  Because the dataset included negative values, we also, prior to sensor scaling, subtracted the minimum values of each feature (as obtained from the validation set) to render most feature values positive; any remaining negative data points were clipped to zero.  To better assess online learning, we split the dataset into 4 groups corresponding to the four forest type classes, and trained with each of these groups in sequence: Sugi  (\textit{group 1}), Hinoki (\textit{group 2}), Mixed deciduous (\textit{group 3}), Other (\textit{group 4}).  

\subsection{Species-specific anuran call dataset}
This dataset includes acoustic features (mel frequency cepstral coefficients, MFCCs) extracted from the call syllables of 10 different frog and toad species, recorded in the wild in Brazil and Argentina \cite{colonna_incremental_2015, colonna_recognizing_2016, colonna_automatic_2016, colonna_how_2016, ribas_similarity_2012, diaz_compressive_2012, colonna_distributed_2014, colonna_feature_2012, noauthor_uci_nodate, ucirepo}. The dataset includes 7195 samples, with each sample comprising 22 MFCC features (values between $-1$ and $1$), and exhibits significant class imbalance; i.e., the numbers of samples corresponding to each class (species) differ substantially. To make all data samples positive, we shifted each value by $+1$ so that each MFCC feature was in the range $0$ - $2$. 

This dataset, uniquely among those tested, also included multilabel, multiclass classification, enabling us to illustrate the algorithm's innate capacity for natural hierarchical representation. Specifically, while training was performed using only species information (10 groups), we also measured the classification of calls into the correct anuran genus and family. Altogether, the 10 species in the dataset comprise 8 anuran genera within 4 families. $10\%$ of the data were retained as a validation set, although these data were not used because the feature range was already known to be between $0$ to $2$ and hence validation per se was not required.  As above, to assess online learning, we split the dataset into 10 groups corresponding to the 10 species, and trained with each in series: \textit{Adenomera andre} (family Leptodactylidae, \textit{group 1}), \textit{Adenomera hylaedactylus} (family Leptodactylidae, \textit{group 2}), \textit{Ameerega trivittata} (family Dendrobatidae, \textit{group 3}), \textit{Hyla minuta} (since reclassified as \textit{Dendropsophus minutus}, family Hylidae, \textit{group 4}), \textit{Hypsiboas cinerascens} (family Hylidae, \textit{group 5}), \textit{Hypsiboas cordobae} (family Hylidae, \textit{group 6}), \textit{Leptodactylus fuscus} (family Leptodactylidae, \textit{group 7}), \textit{Osteocephalus oophagus} (family Hylidae, \textit{group 8}), \textit{Rhinella granulosa} (family Bufonidae, \textit{group 9}), \textit{Scinax ruber} (family Hylidae, \textit{group 10}). 

\section{Empirical results}
After training the network with standard heterogeneous parameters, and tuning the  $w_{max}$ distribution on the validation set of batch 1 of the UCSD chemosensor drift dataset, we trained the algorithm and tested its performance on three different datasets as described above.  Specifically, we measured (1) the goodness of preprocessing ($g_{p}$) for each dataset, to assess how well the same instantiated, parameterized network would operate across a statistically diverse range of inputs, and (2) an interim estimate of classification performance based on a thresholded Hamming distance between activated ensembles in the interneuron representation, omitting the recurrent feedback loop of the full model (Figure ~\ref{schematic}).  The latter measure is reported in order to illustrate the importance of signal conditioning (Table \ref{tab:perf.}), and should not be used as a benchmark for the performance of the intact algorithm, which  classifies signals successfully under high levels of synthetic impulse noise \cite{imam_rapid_2019}.  

\begin{figure}[ht]
  \centering
  \includegraphics[width=2in]{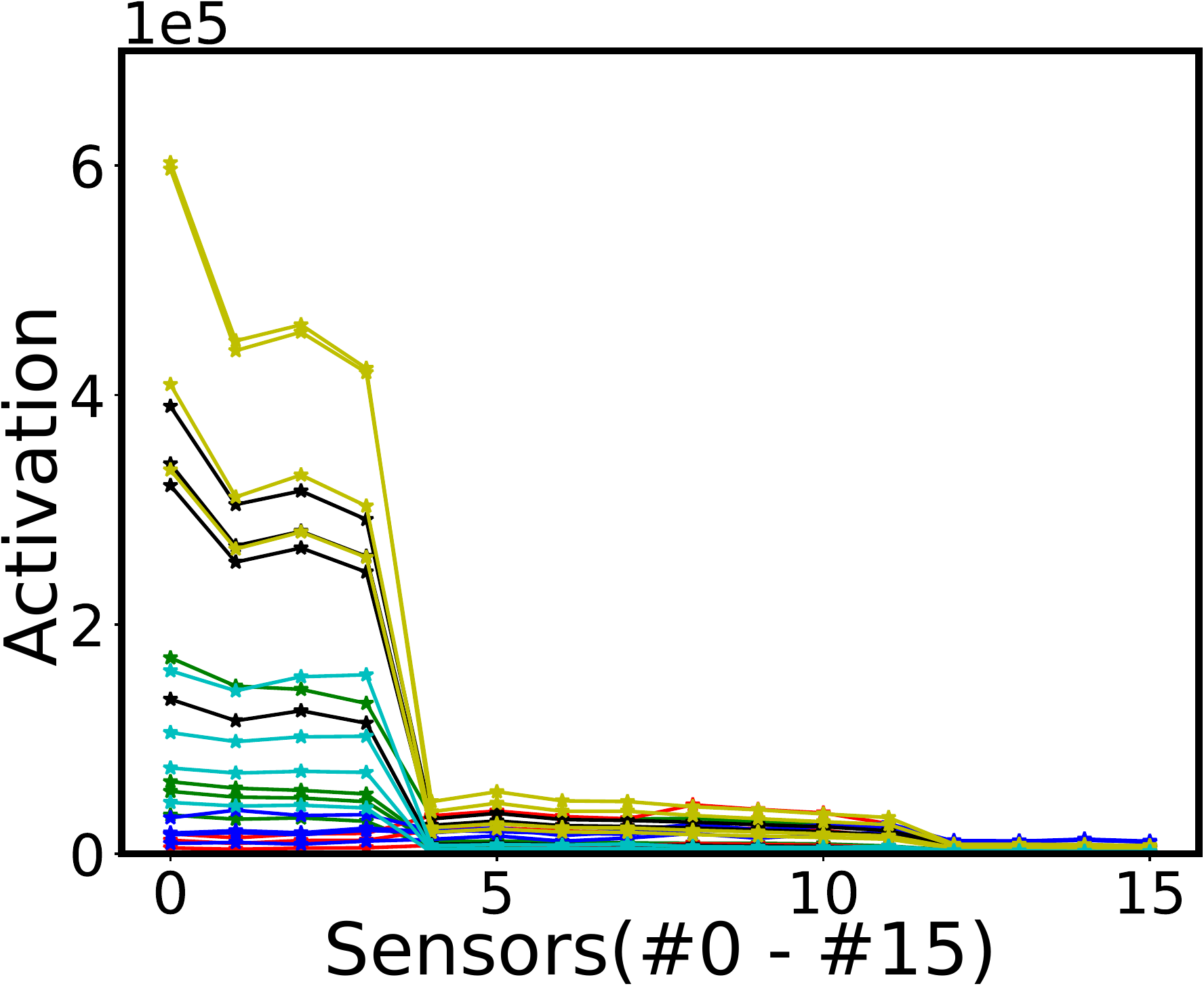}
  \caption{Raw sensor data from Batch 1 of the UCSD chemosensor drift dataset.  Four samples of each of 6 different odorants, presented at different, randomly selected, concentrations, are depicted. The mean response amplitudes of sensors differ widely across both samples and sensors.}  
  \label{raw_b1}
\end{figure}

 More important for present purposes is the uniformity of interneuron recruitment levels across a statistically diverse set of raw input signals, as assessed by $g_{p}$.  Direct inputs from deployed sensors differ substantially (Figure \ref{raw_b1}).  As the distribution of response amplitudes across a sensor array strongly affects the efficacy of interneuron recruitment in this framework, and interneuron recruitment profiles substantially determine learning and classification performance, input patterns consequently must be transformed to exhibit a relatively consistent statistical structure in order to avoid the need to retune network parameters, and hence enable learning in the wild.

\subsection{Preprocessors}
To assess preprocessor efficacy, we first implemented a 16-column network including 16 principal neurons (MCs) and $3200$ inhibitory interneurons (GCs), and presented this network with Batch 1 data from the UCSD sensor drift dataset \cite{vergara_chemical_2012, rodriguez-lujan_calibration_2014}. Table ~\ref{tab:drift_gp} depicts the goodness of preprocessing ($g_{p}$; Equation ~\ref{gp}) for each of the sequential preprocessing steps, and Figure \ref{b1_pre} illustrates the successive transformations of sensor input distributions and the uniformity of interneuron recruitment following each preprocessing step. For example, Figure ~\ref{b1_pre}a depicts the raw sensor responses to 24 odorant presentations sorted by amplitudes. Interneuron recruitment into the active ensemble by these raw sensor inputs (after being linearly scaled by a factor of \num{5e-5}) differed substantially among samples and was zero for some lower-concentration samples, resulting in a $g_{p}$ value of zero (Table ~\ref{tab:drift_gp}). Subsequent preprocessor stages regularize the distribution of input amplitudes and improve interneuron recruitment uniformity ($g_{p}$; Table ~\ref{tab:drift_gp}).

\subsubsection{Sensor scaling:}  Heterogeneous sensor arrays require sensor-specific rescaling to a common range so that sensors producing the largest output ranges do not inappropriately dominate network operations. Accordingly, in the first preprocessing step, we scaled both the training set and the test set by the maximum observed sensor responses determined from the  $10\%$ validation set of Batch 1 (uniform sensor scaling). We then further scaled all inputs by a equidimensional uniform vector $v_{uni}$, where $v_{uni} \in [0.5, 1.0]$ (nonuniform sensor scaling). Sensor response profiles now become more comparable in amplitude, but still exhibit concentration-dependent activation profiles (Figure \ref{b1_pre}b) and, if anything, exhibit less uniform interneuron recruitment (Figure \ref{b1_pre}f; Table ~\ref{tab:drift_gp}). 

\subsubsection{Unsupervised intensity normalization:}  Distinguishing concentration differences from genuine quality differences in the biological system (concentration tolerance) depends in part on a global inhibitory feedback mechanism instantiated in the MOB glomerular layer \cite{cleland_sequential_2011, banerjee_interglomerular_2015}.  We applied this normalization operation to the output of the sensor scaling preprocessor. The diverse sensor response profiles observed for the same gas types in Figure \ref{b1_pre}b arise from concentration differences; this preprocessor substantially eliminates those within-type differences (Figure \ref{b1_pre}c,g; Table ~\ref{tab:drift_gp}). Notably, this step removes the need to train the algorithm with multiple concentrations of a given gas type, enabling \textit{generalization beyond experience} in the concentration domain. 

\subsubsection{Heterogeneous duplication:}  
\begin{figure*}[ht]
\centering
\subfloat[Raw data]{\includegraphics[width=0.24\linewidth]{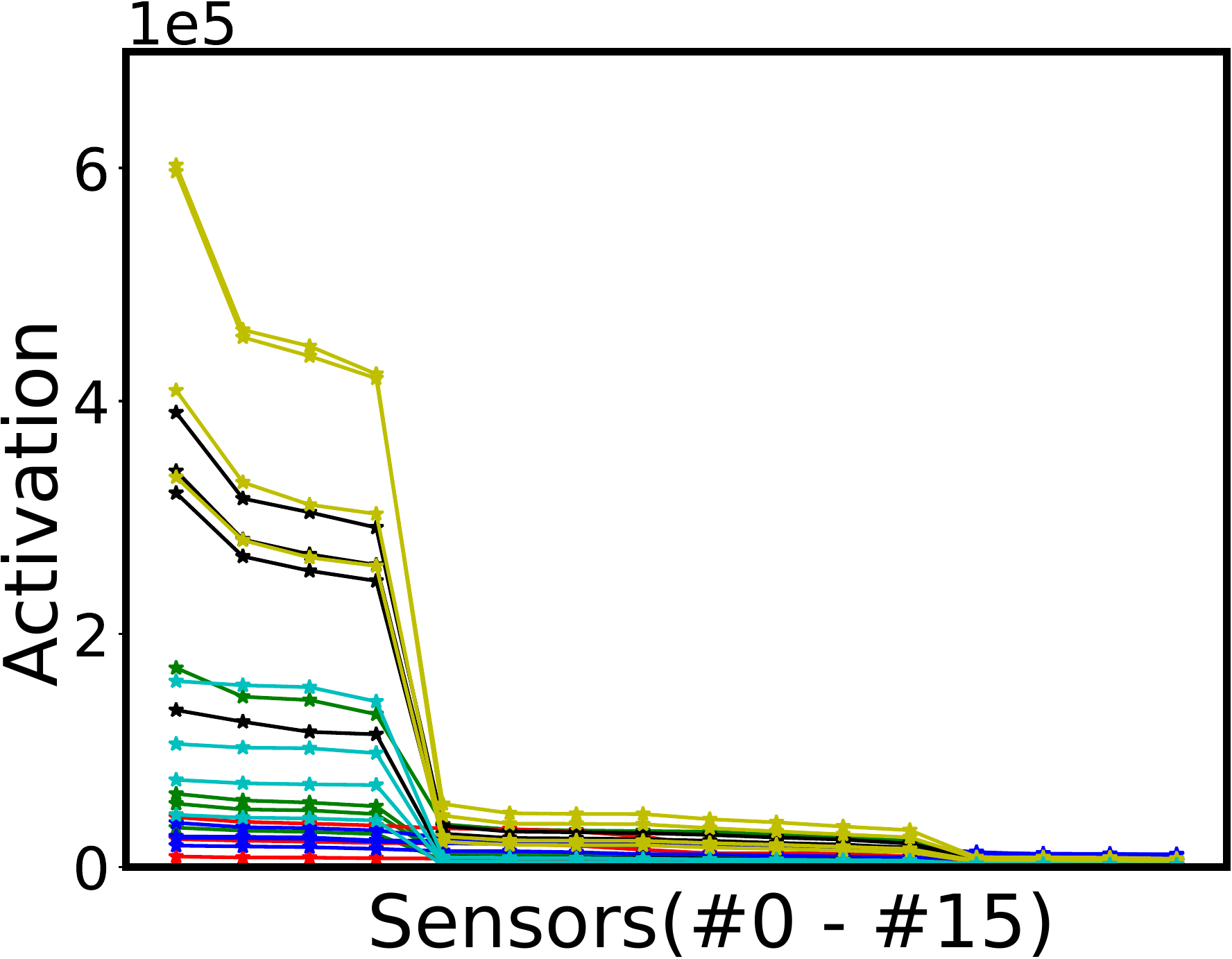}}
\hspace{0.00005in}
\subfloat[Sensor scaling]{\includegraphics[width=0.24\linewidth]{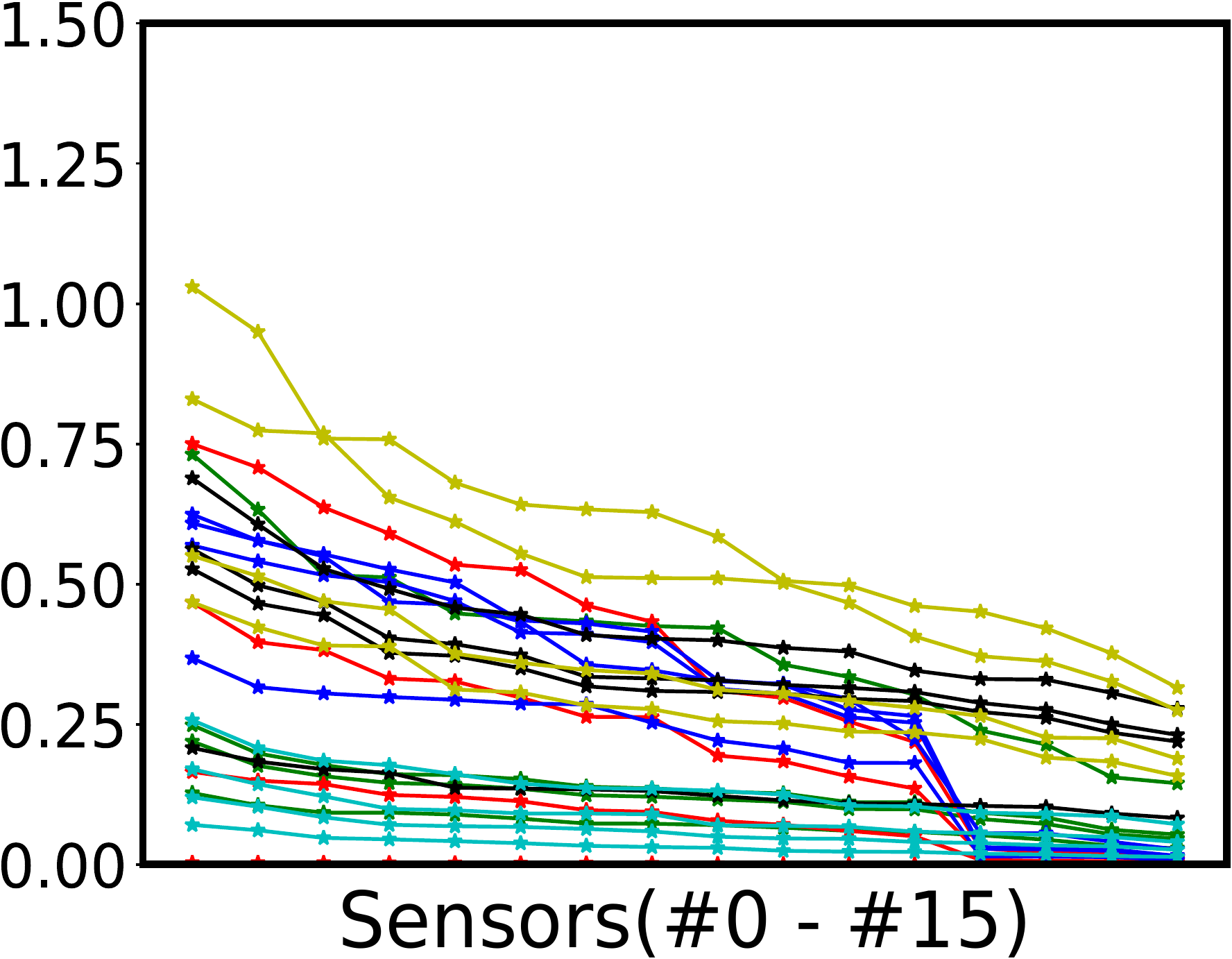}}
\hspace{0.00005in}
\subfloat[Intensity normalization]{\includegraphics[width=0.24\linewidth]{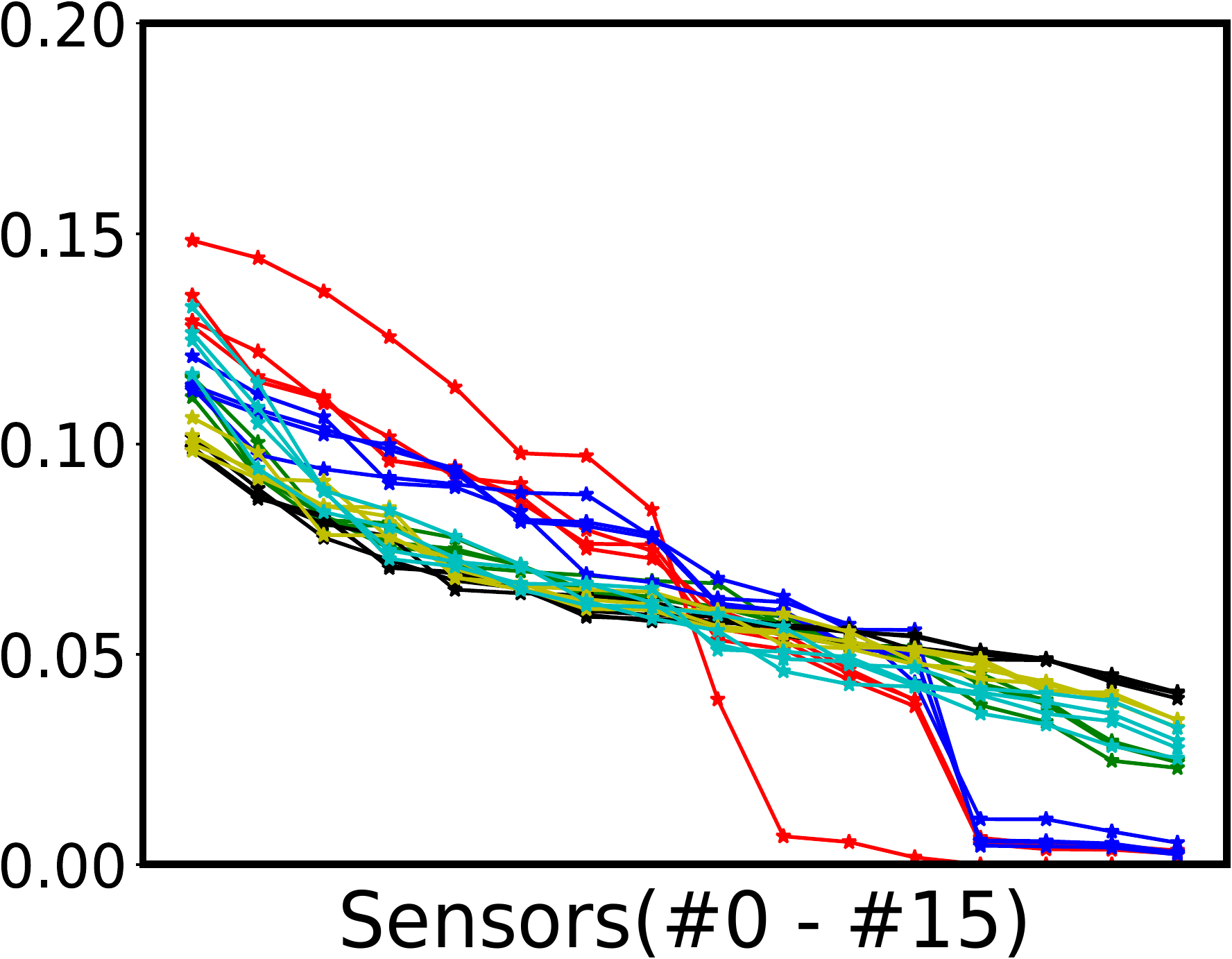}}
\hspace{0.00005in}
\subfloat[Heterogeneous duplication]{\includegraphics[width=0.24\linewidth]{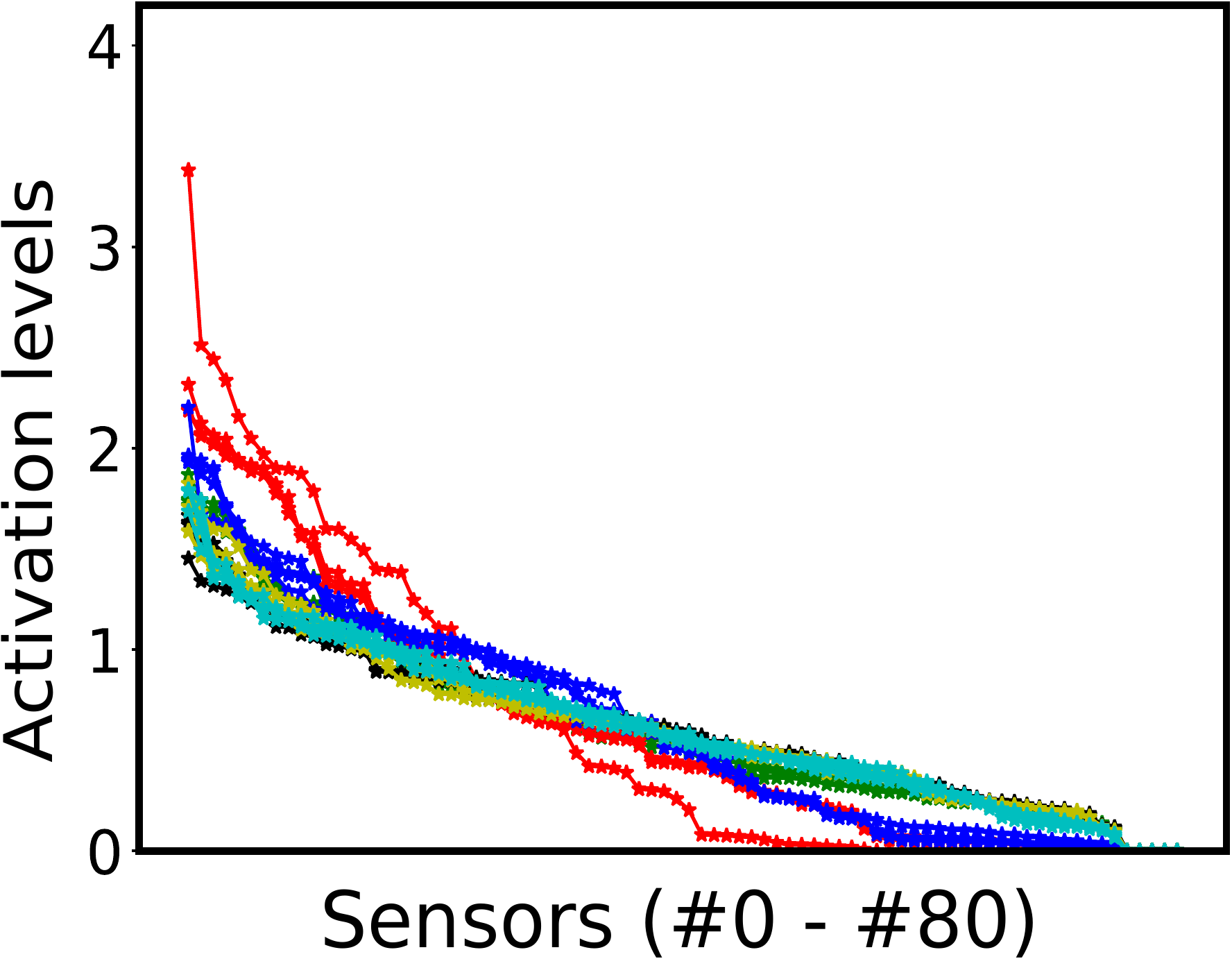}}

\subfloat[]{\includegraphics[width=0.24\linewidth]{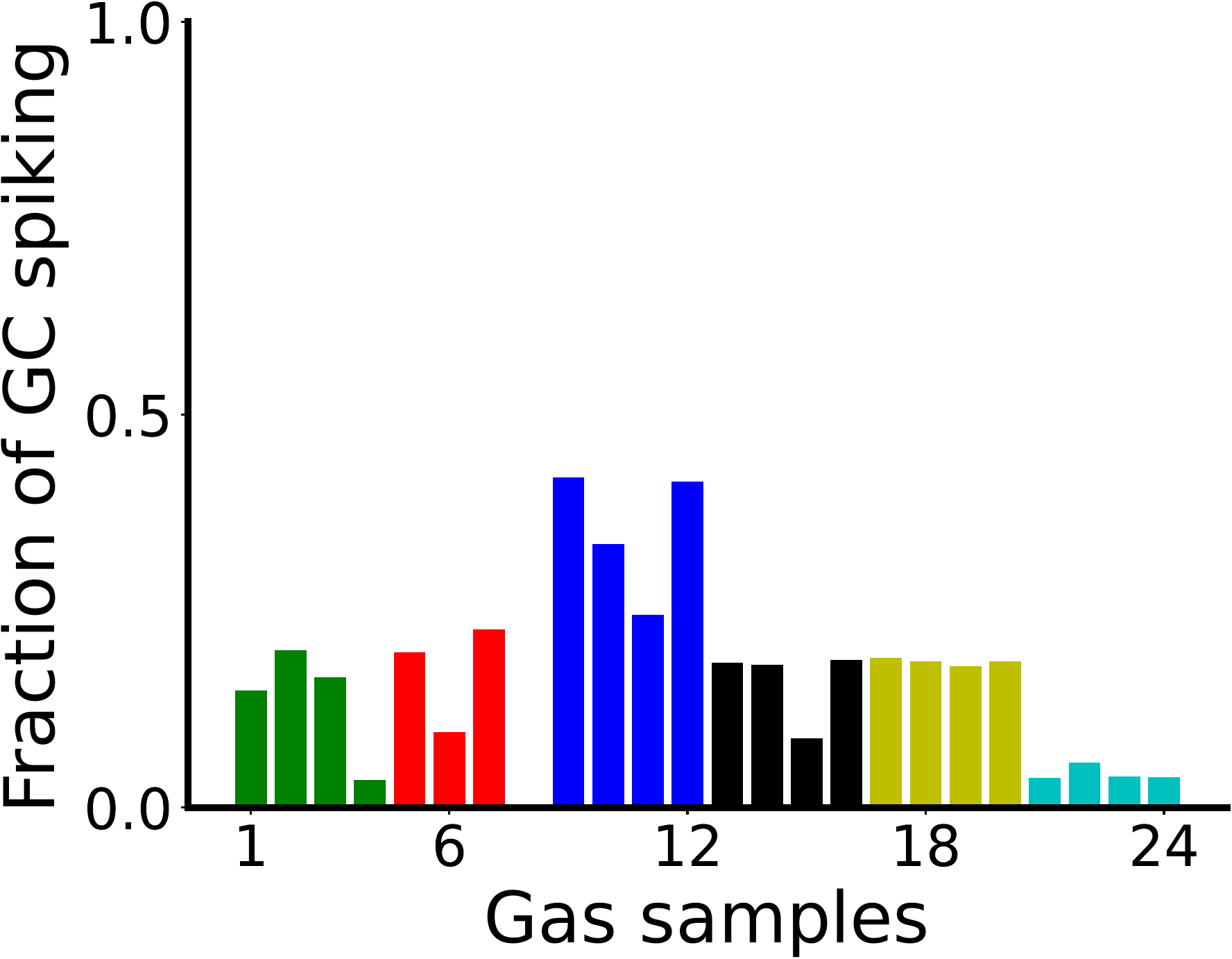}}
\hspace{0.00005in}
\subfloat[]{\includegraphics[width=0.24\linewidth]{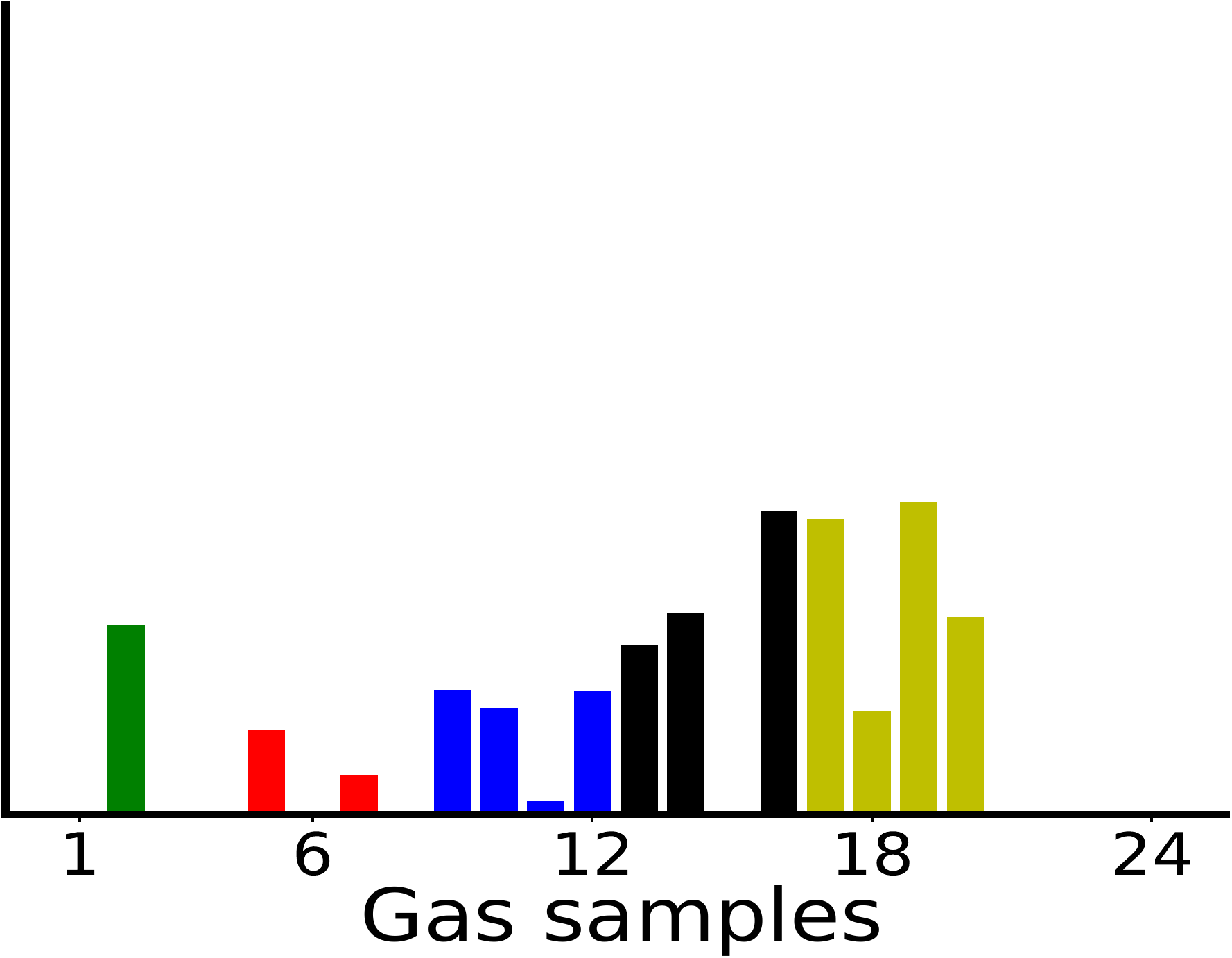}}
\hspace{0.00005in}
\subfloat[]{\includegraphics[width=0.24\linewidth]{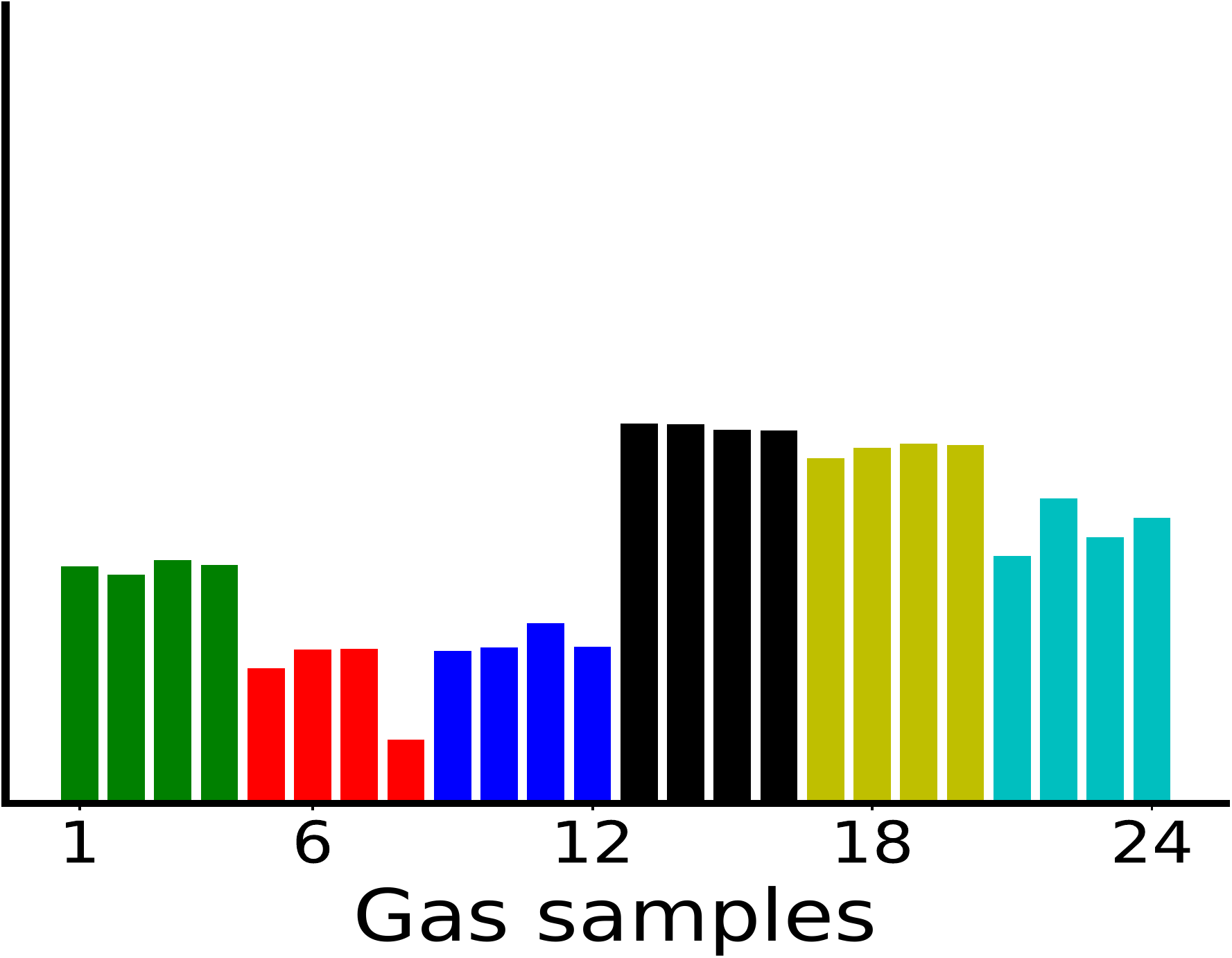}}
\hspace{0.00005in}
\subfloat[]{\includegraphics[width=0.24\linewidth]{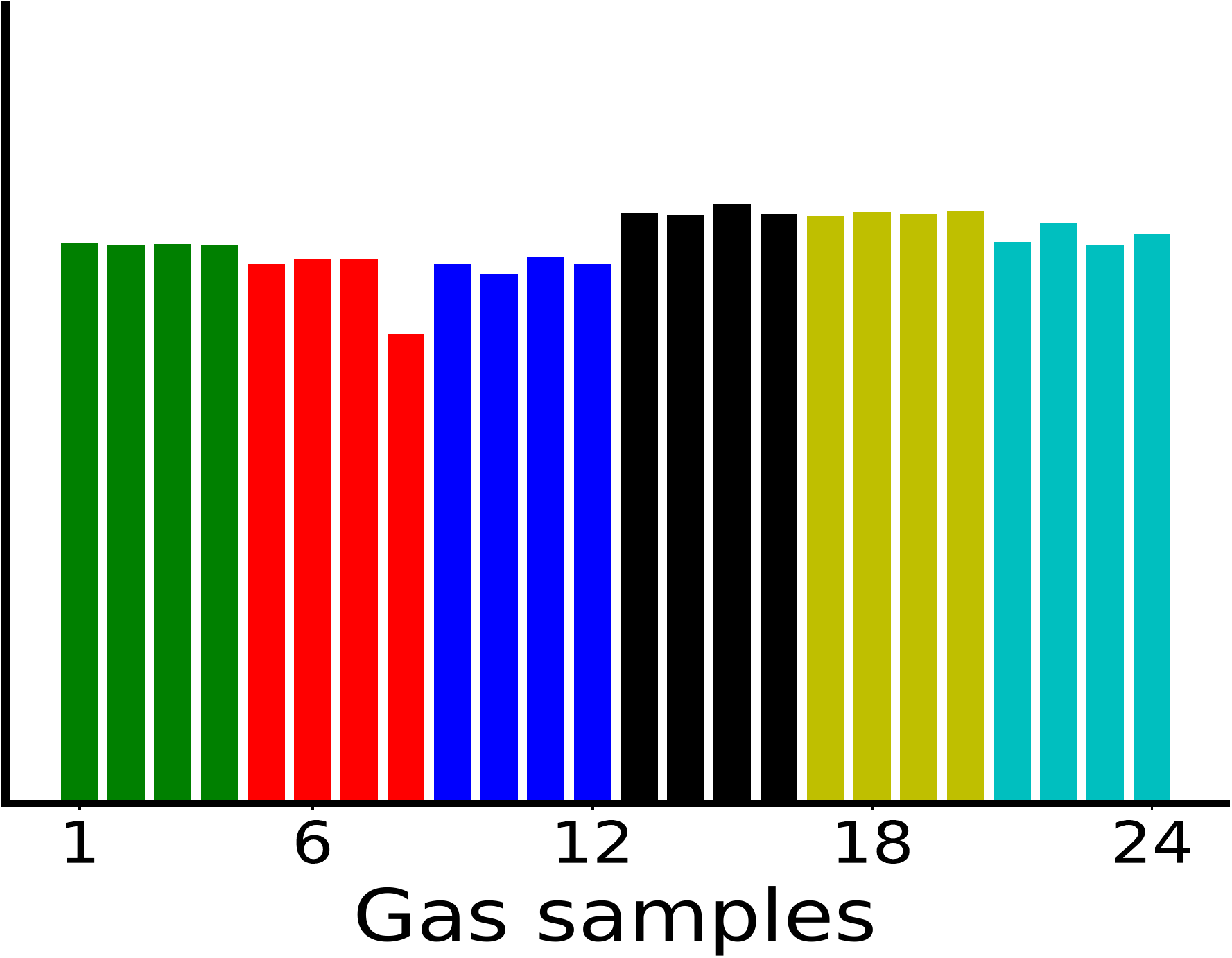}}
\caption{Application of successive preprocessing steps to four random samples, with different concentrations, drawn from each of the 6 gas types from Batch 1 of the UCSD gas sensor drift dataset. (a-d) Distributions of activation levels across the sixteen sensory inputs in the dataset, measured at the input to the principal neurons (MCs) of the core network.  In panel (d), heterogeneous duplication has increased the number of MCs by a factor of five.  Each successive step transforms the sensory input distribution until it is substantially regularized. (e-h) The proportion of the interneuron (GC) population that is activated by each sample differs from sample to sample in the early stages of signal conditioning; after signal conditioning is complete (panel h), statistically diverse inputs, across odorant types and concentrations, recruit effectively uniform numbers of activated interneurons.}
\label{b1_pre}
\end{figure*}

In this step, the output of the intensity normalization preprocessor first is projected to a higher dimension in a column-specific manner by duplicating each output onto $m$ feedforward excitatory interneurons with heterogeneous properties and then randomly connecting those interneurons to $n$ principal neurons (MCs), thereby multiplying the number of columns of the subsequent core learning network by a factor of $n$ (column duplication).  In the present simulations, $m$ = $n$ = 5 (Figure~\ref{het_dup}).  After applying this preprocessing step, sensor response distributions become regularized (Figure \ref{b1_pre}d) and interneuron recruitment becomes substantially uniform across samples, exhibiting a $g_{p}$ of 0.94 for Batch 1 data (Figure \ref{b1_pre}h; Table ~\ref{tab:drift_gp}). Importantly, this transformation occurs in a naturally online manner, without destroying inherent similarity relationships among data samples or reducing test set classification performance. 

These sequential preprocessor steps, which we refer to collectively as signal conditioning, ensure that statistically diverse inputs are transformed so as to recruit comparable numbers of interneurons, and consequently can be effectively learned and classified by the same instantiated, parameterized network.

\begin{table}
  \caption{$g_{p}$ values assessed from each experimental dataset following the sequential application of each preprocessor.}
  \label{tab:drift_gp}
  \begin{tabular}{ccccc}
    \toprule
     & Raw & Scaled & Intensity norm. & Het. duplication\\
    \midrule
    Batch 1 & 0 & 0 & 0.67 & \textbf{0.94}\\
    \midrule
    Batch 7 & 0 & 0 & 0.80 & \textbf{0.95}\\
    \midrule
    Forest & 0 & 0.85 & 0.90 & \textbf{0.97}\\
    \midrule
    Anuran & 0 & 0.92 & 0.93 & \textbf{0.99}\\
  \bottomrule
\end{tabular}
\end{table}

\begin{figure*}
  \centering
  \subfloat[]{\includegraphics[width=0.28\linewidth]{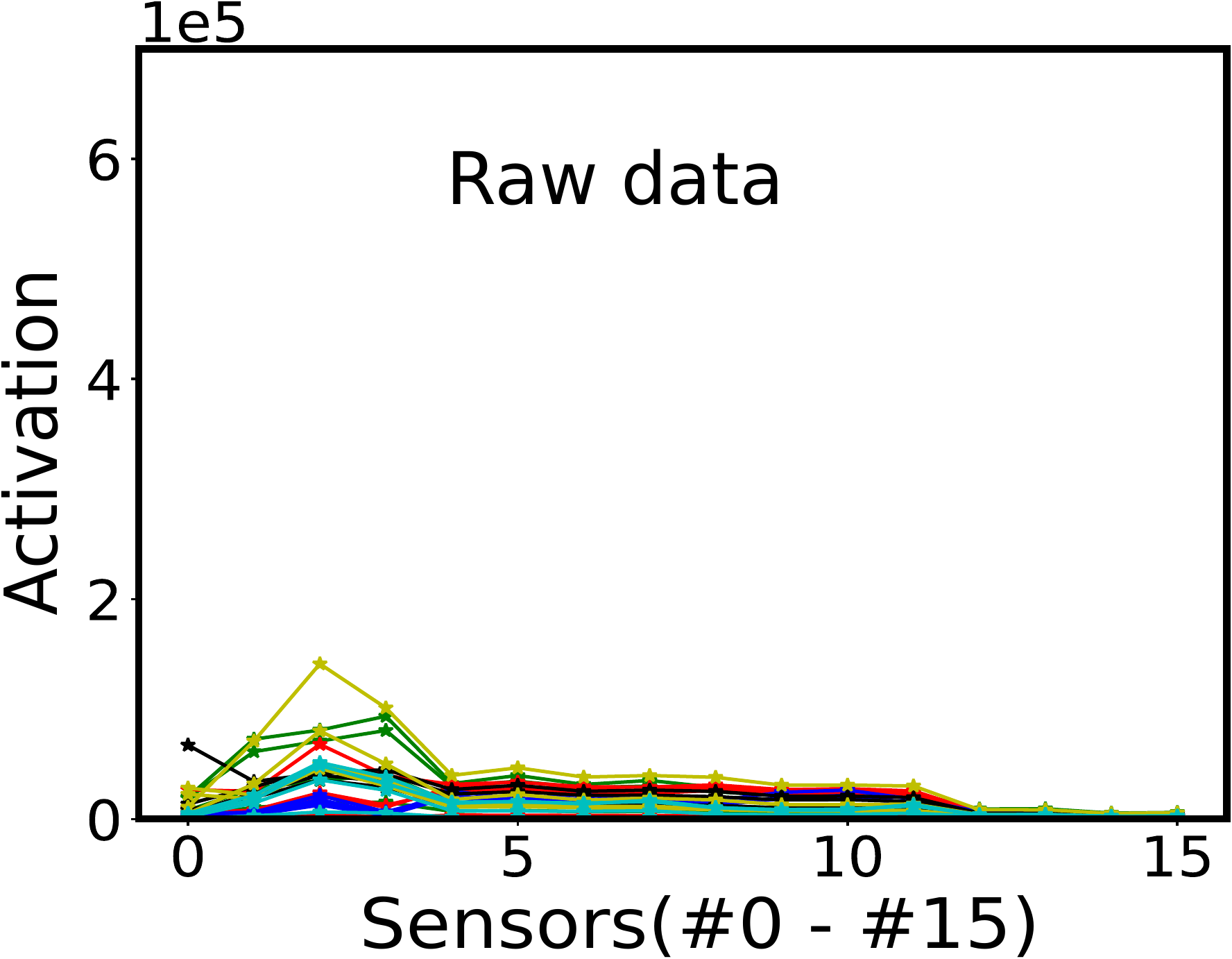}}
    \hspace{0.25in}
\subfloat[]{\includegraphics[width=0.28\linewidth]{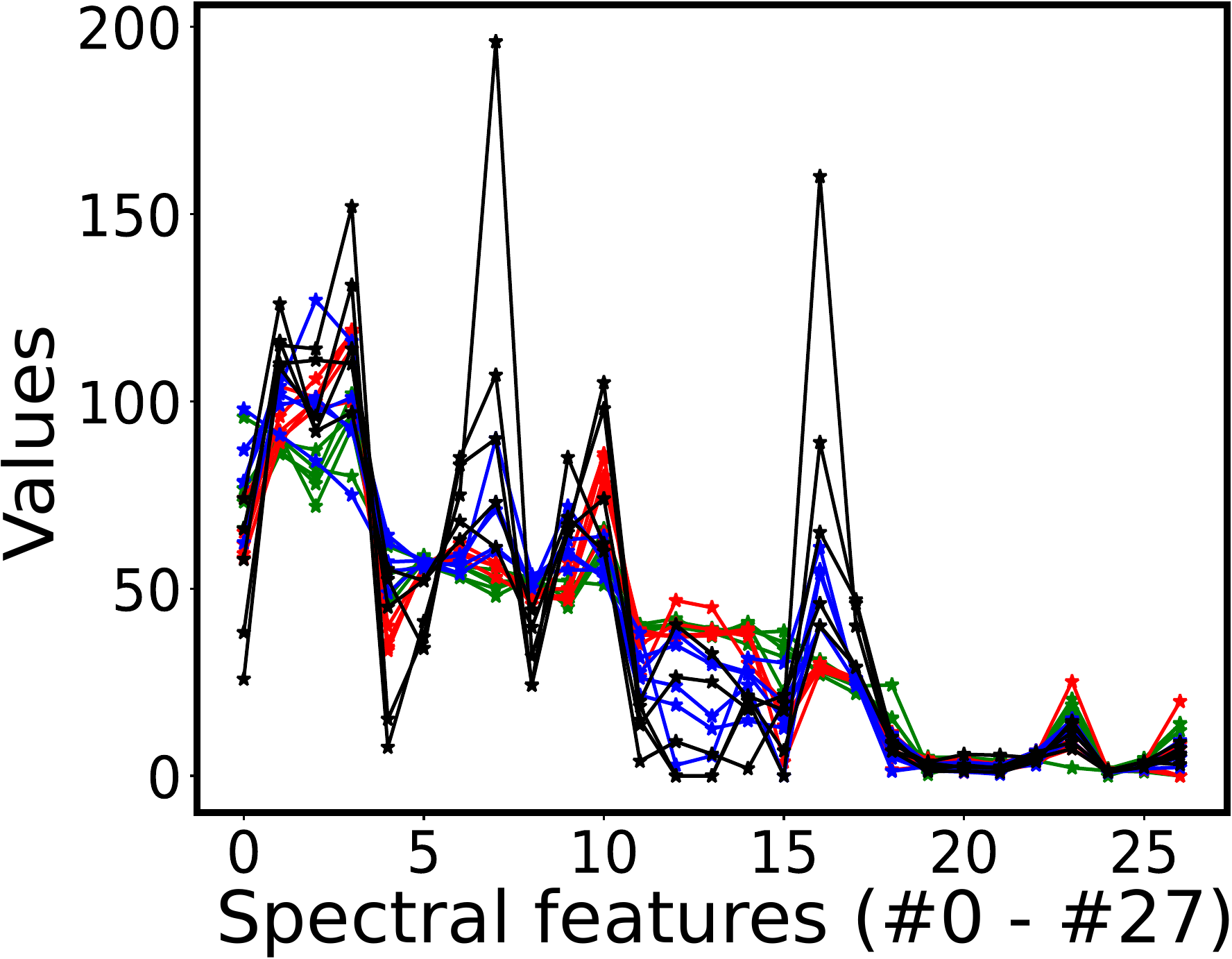}}
\hspace{0.25in}
\subfloat[]{\includegraphics[width=0.28\linewidth]{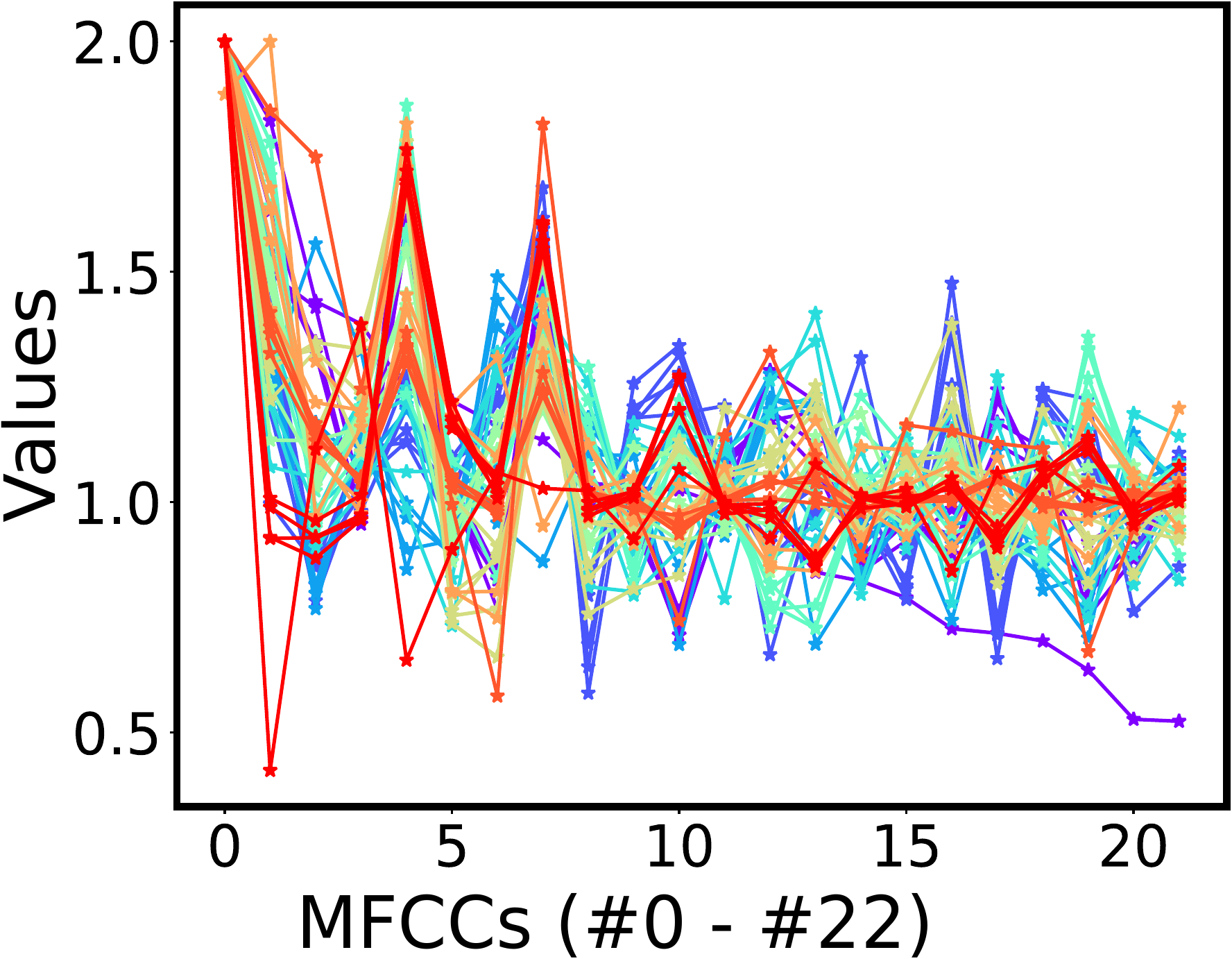}}
\hspace{0.25in}

\subfloat[]{\includegraphics[width=0.28\linewidth]{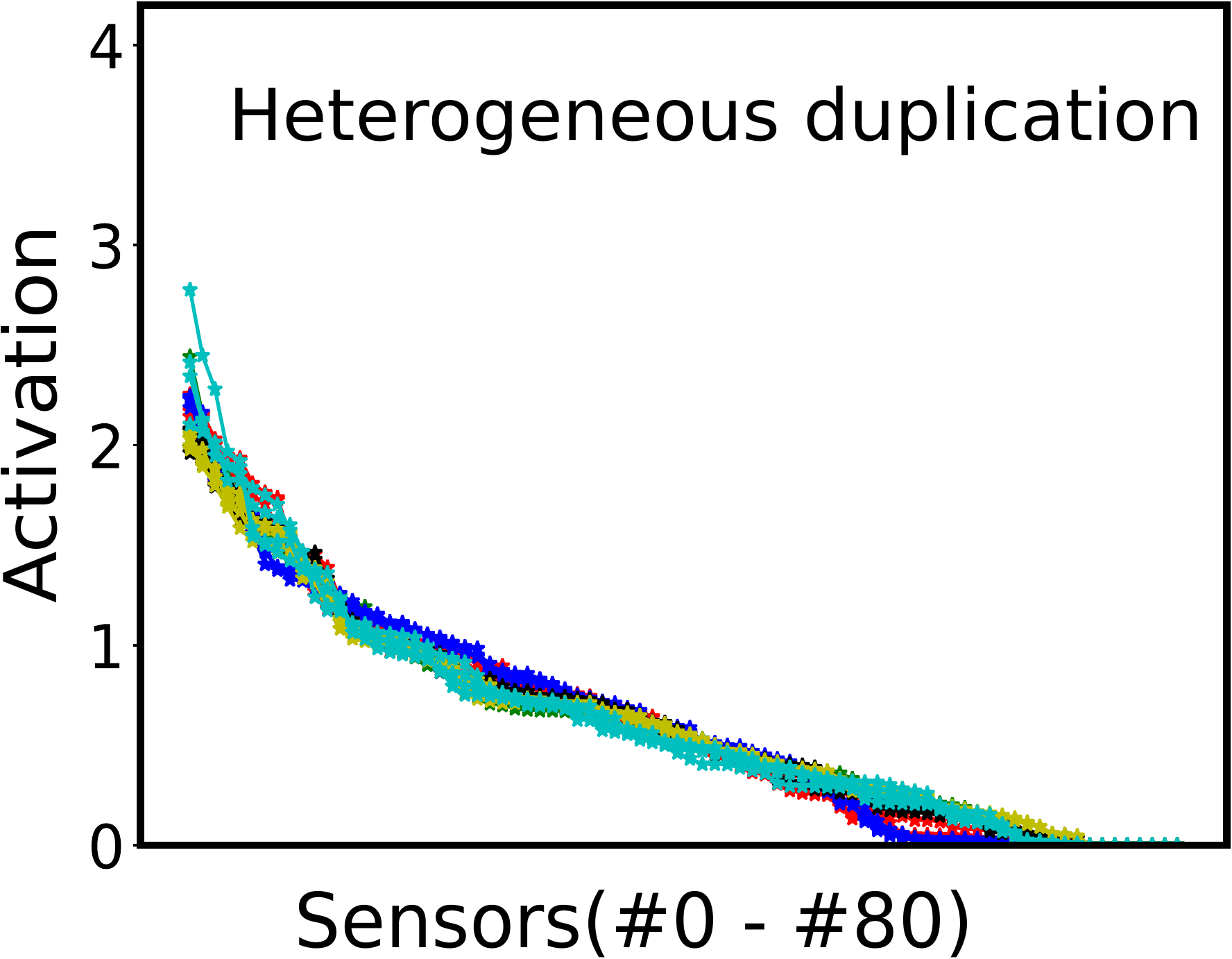}}
\hspace{0.25in}
\subfloat[]{\includegraphics[width=0.28\linewidth]{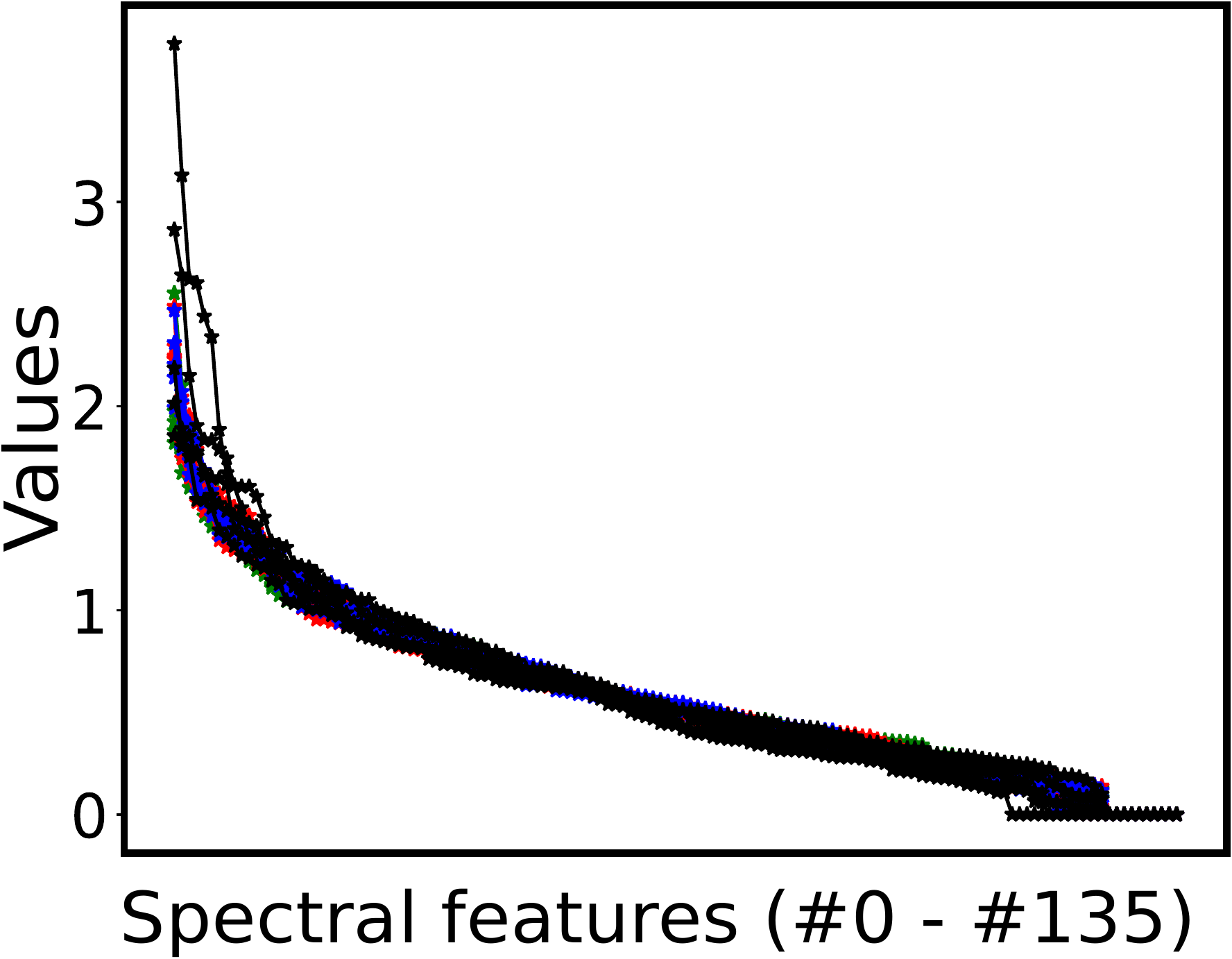}}
\hspace{0.25in}
\subfloat[]{\includegraphics[width=0.28\linewidth]{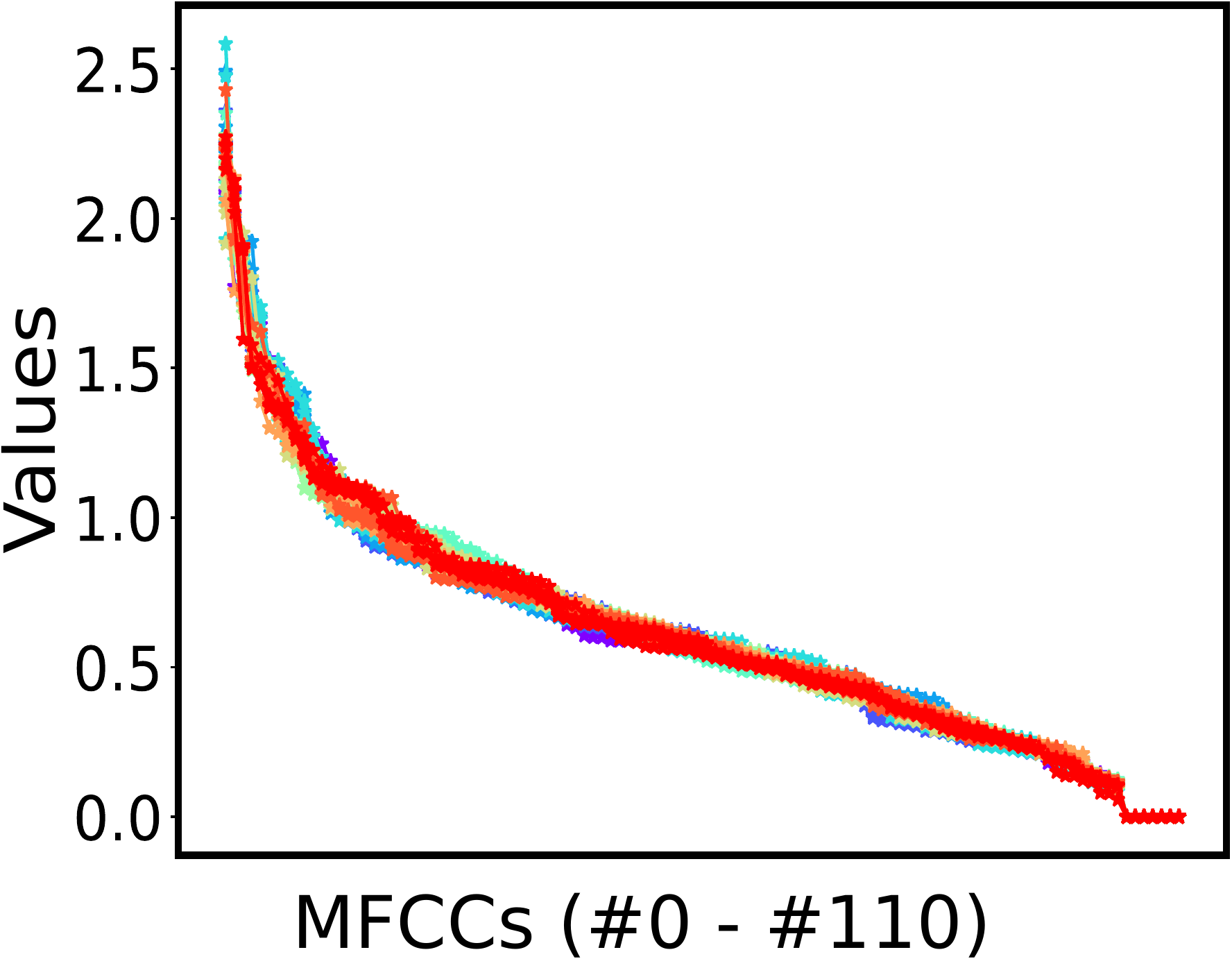}}
\hspace{0.25in}

\subfloat[]{\includegraphics[width=0.28\linewidth]{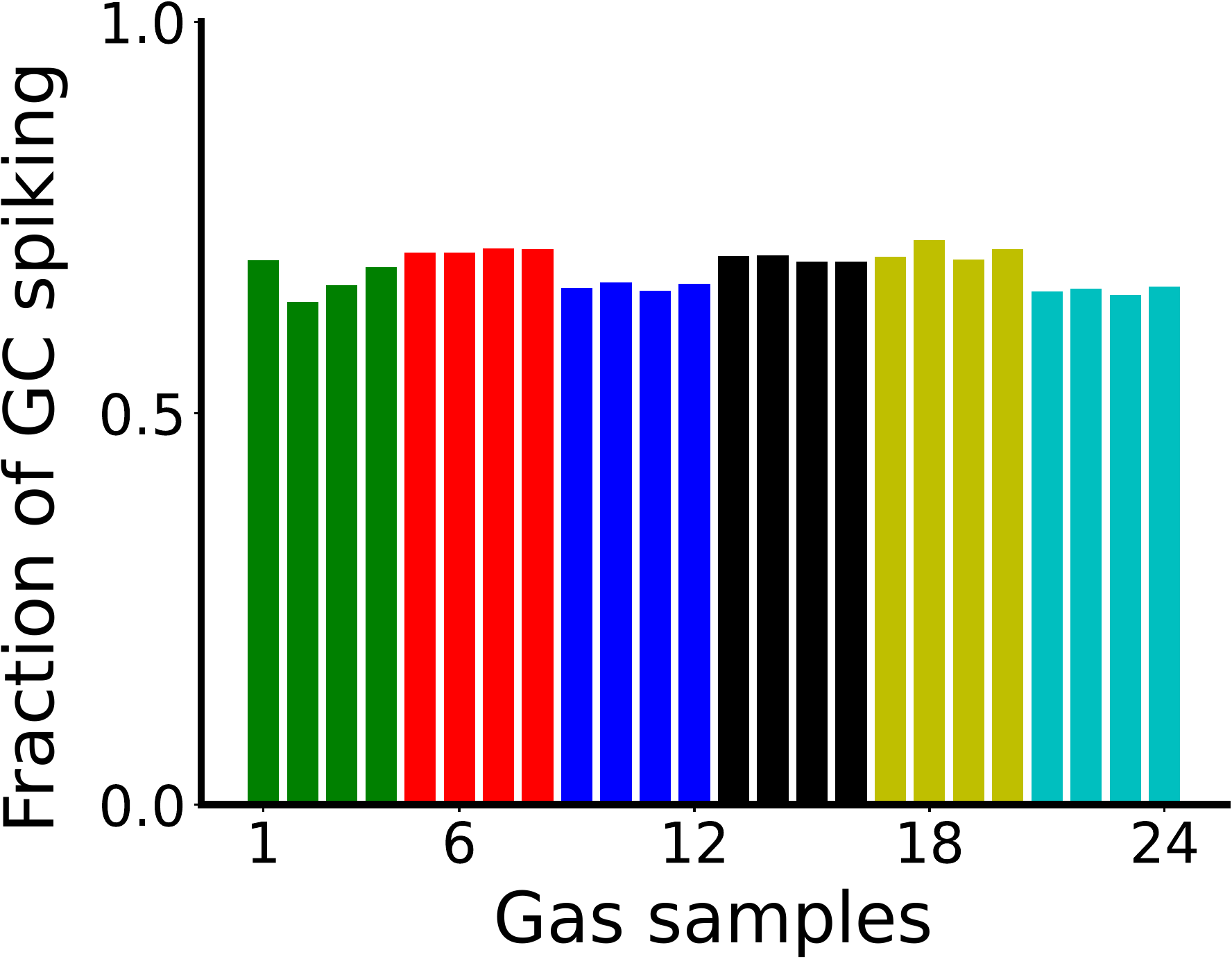}}
\hspace{0.25in}
\subfloat[]{\includegraphics[width=0.28\linewidth]{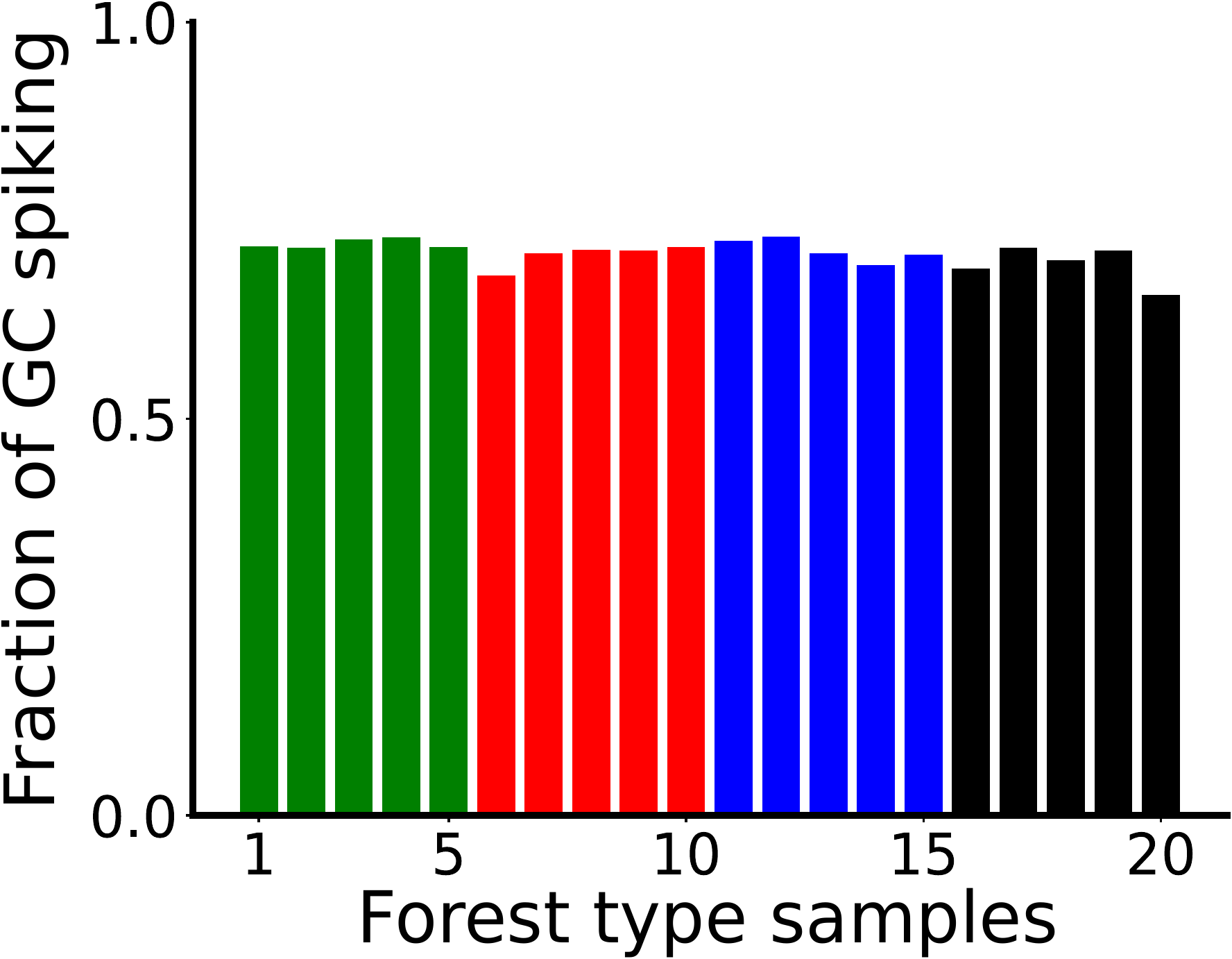}}
\hspace{0.25in}
\subfloat[]{\includegraphics[width=0.28\linewidth]{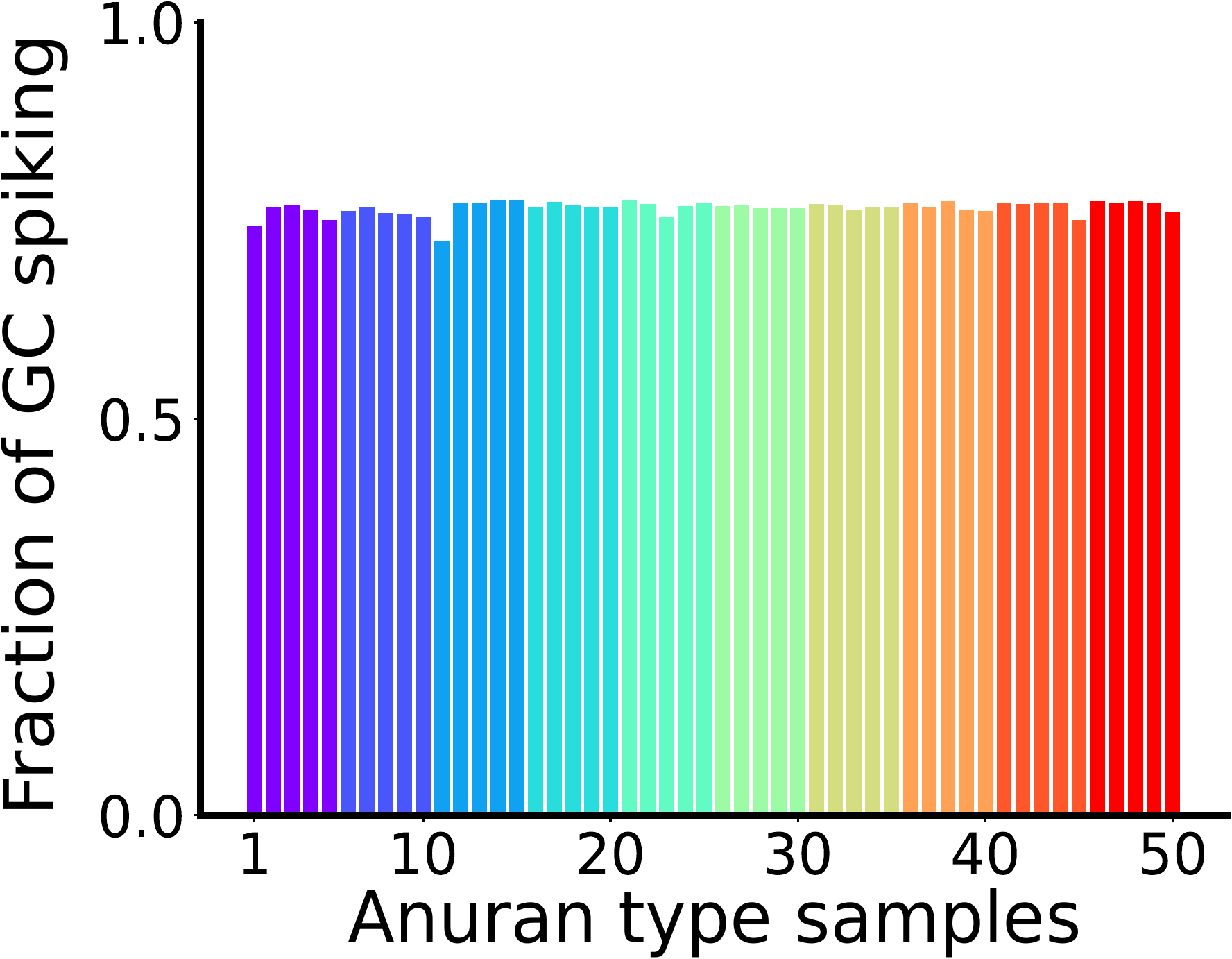}}
\hspace{0.25in}
  \caption{Preprocessor-based transformations of (a, d, g) Batch 7 of the UCSD sensor drift dataset (four samples from each of six gas types), (b, e, h) the forest type spectral mapping dataset (five samples from each of 4 types), and (c, f, i) the anuran call classification dataset (five samples from each of 10 frog or toad species).  (a-c) The raw data are statistically diverse across datasets. (d-f) Signal conditioning via the preprocessor cascade renders each of these datasets statistically consistent with one another and with Batch 1 data (Figure \ref{b1_pre}d).  (g-i) Accordingly, all samples recruit similar numbers of interneurons into the active ensemble ($g_{p}$ = 0.95, 0.97, 0.99 respectively). Note that the dimensionalities of these datasets also differ (16, 27, and 22 dimensions respectively).}  
  \label{all_pre}
\end{figure*}

\subsection{UCSD gas sensor drift}
\begin{figure}
  \centering
  \subfloat[]{\includegraphics[width=0.45\linewidth]{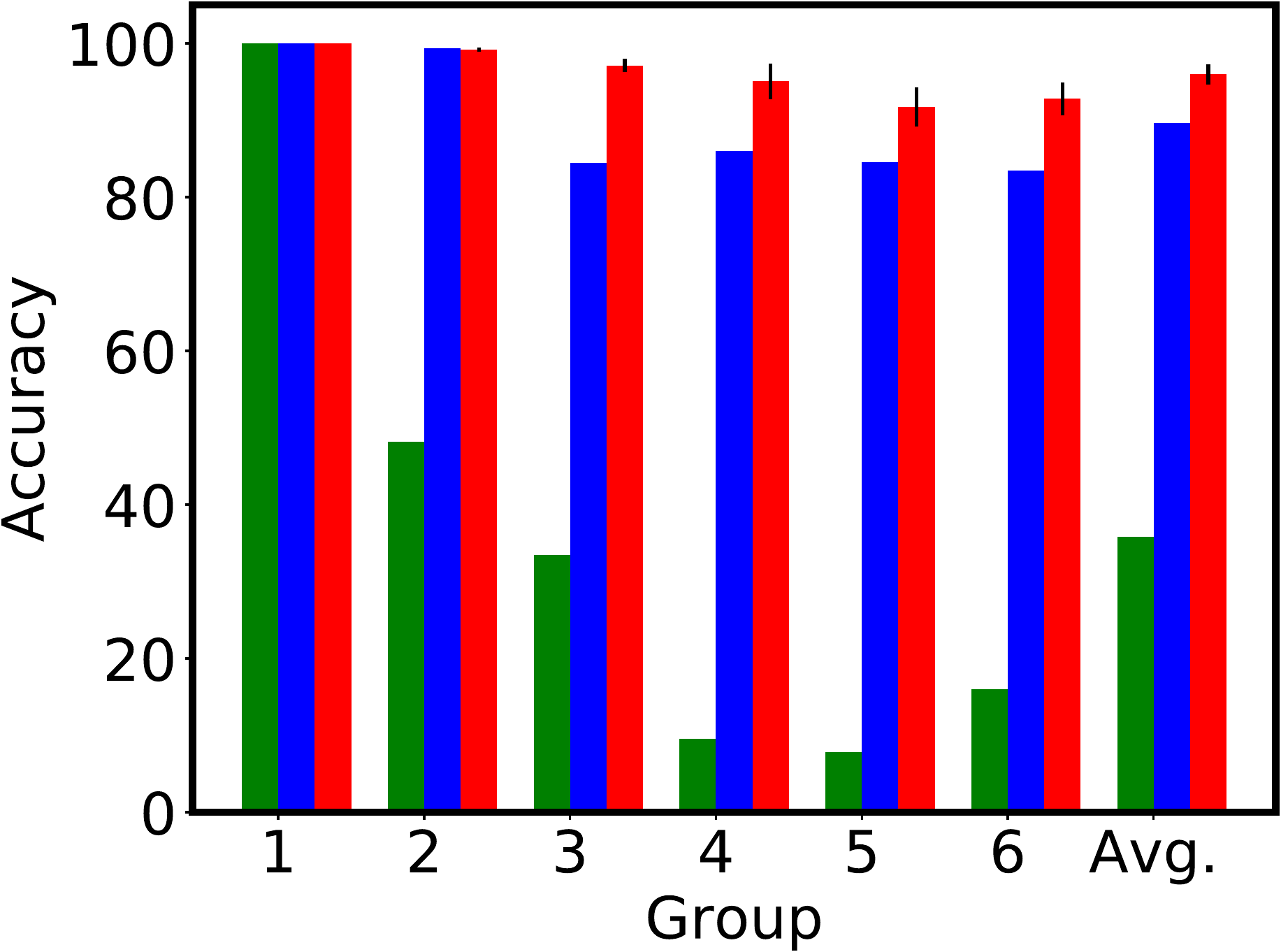}}
    \hspace{0.25in}
    \subfloat[]{\includegraphics[width=0.45\linewidth]{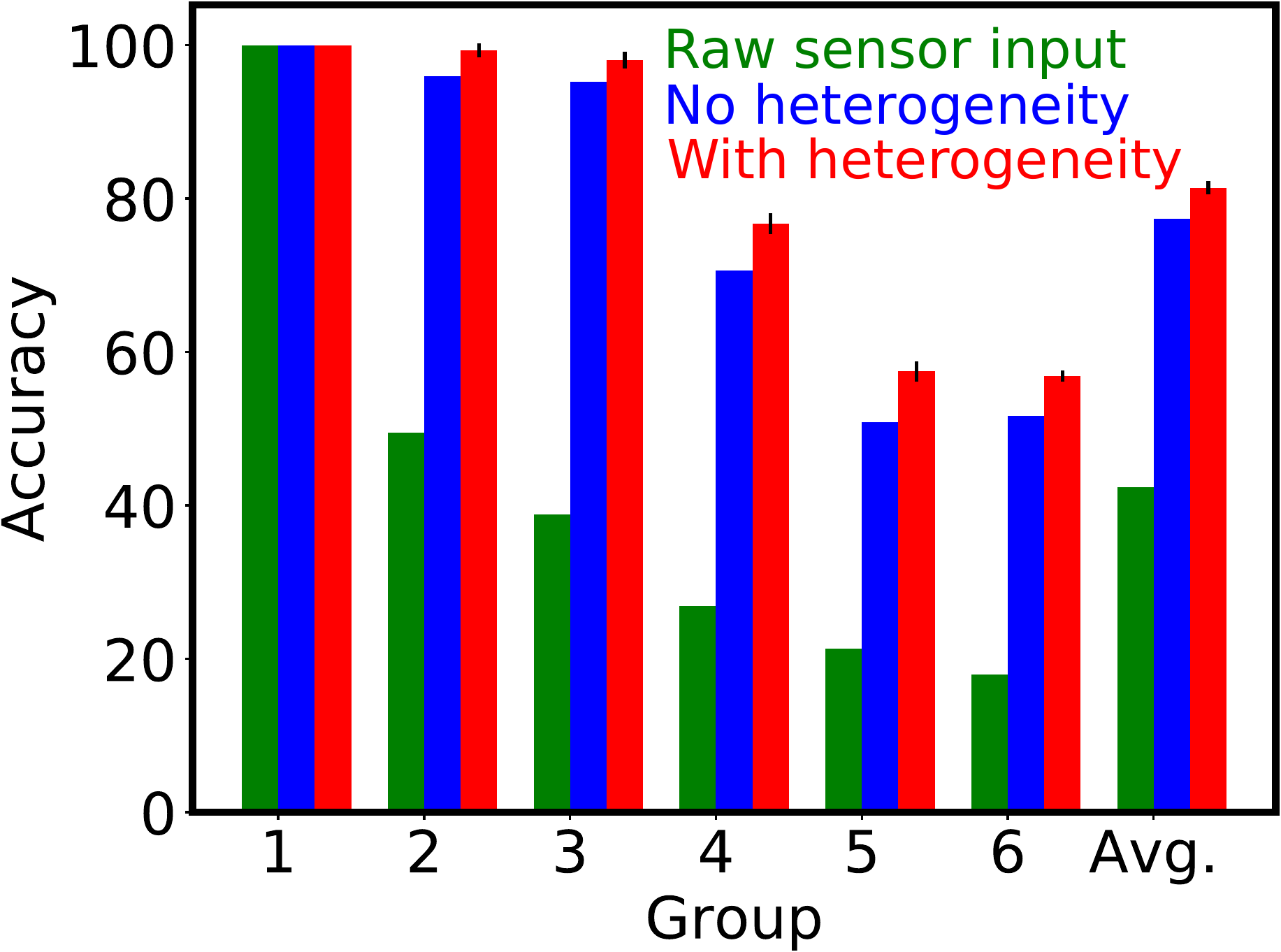}}
  \caption{Comparisons of classification performance in an online learning setting using (a) Batch 1 and (b) Batch 7 data from the UCSD sensor drift dataset.  Performance was compared across three conditions: when raw sensor input was directly provided to the learning network (\textit{green bars}), when signal conditioning was performed without parameter heterogeneity (\textit{blue bars}), and when all preprocessors and parameter heterogeneities were fully operational (\textit{red bars}). \textit{Avg.} depicts classification performance averaged across the six individual assessments.}
  \label{drift_het17}
\end{figure}

Using these preprocessors, we tested the \textit{learning in the wild} capability of our feedforward learning network, first using Batch 1 data, and then, without changing any network parameters, Batch 7 data. We first trained the network on raw sensor data from Batch 1 using one-shot learning with odorant concentration uncontrolled. In total, the training set constituted $1.35\%$ of the dataset. As noted above, we trained on each group (odorant type) in sequence, testing performance on all six groups at each step (with odorants from untrained groups generating \textit{none of the above} classifications). Unsurprisingly, performance deteriorated after training on two or more groups, with the average accuracy across all training stages being only $35.86\%$ (Figure ~\ref{drift_het17}, \textit{green bars}). Following the same training procedure, but using a network incorporating the preprocessors and heterogeneities described above, we obtained a mean classification accuracy of $96.00\%$ (Table~\ref{tab:perf.}; Figure ~\ref{drift_het17}, \textit{red bars}). 

To assess the effects of heterogeneity per se, we next trained a separate network, using the same parameters and including the three preprocessors, but excluding parameter heterogeneity.  Specifically, this exclusion implied:
\begin{itemize}
    \item No modulation of sensor scaling parameters by an equidimensional random vector.
    \item No heterogeneity in the parameters of feedforward interneurons.
    \item No heterogeneity in core learning network parameters.  
\end{itemize}
In this scenario, the average performance across all 6 groups dropped to $89.66\%$, largely owing to performance reductions in later-trained groups (Figure ~\ref{drift_het17}, \textit{blue bars}). Because of the generally high performance on Batch 1 data, we did not also analyze performance with multiple-shot learning (but see \cite{borthakur_spike_2019}).   

Later batches in the UCSD dataset exhibited responses to odorants that differed sharply from those in earlier batches, owing to gradual sensor contamination and other forms of drift (Figure ~\ref{all_pre}a; compare to Figure ~\ref{b1_pre}a). Because the practical goal of \textit{learning in the wild} is to enable the same instantiated network to operate effectively on statistically diverse datasets, we trained the same network (identical parameters) on these Batch 7 data, which comprise odorant responses from the same sensors as in Batch 1, but following 21 months of sensor degradation \cite{vergara_chemical_2012, rodriguez-lujan_calibration_2014}.  Critically, the sequentially applied preprocessors, with heterogeneity, regularized the distribution of sensor input amplitudes to a form consistent with that of the processed Batch 1 data, resulting in a uniform recruitment of interneurons across samples and concentrations (Figure ~\ref{all_pre}d,g; compare to Figure ~\ref{b1_pre}d,h).  

We trained this network using one-shot learning of randomly selected Batch 7 samples (concentrations uncontrolled), using the same procedures as for Batch 1.  As with Batch 1, performance dropped rapidly as additional groups were learned; the average performance across all stages of learning was $42.42\%$ (Figure~\ref{drift_het17}, \textit{green bars}), with a training set comprising $0.17\%$ of the data.  After applying the three preprocessors, including heterogeneities, average performance improved to $81.42\%$ (Table~\ref{tab:perf.}; Figure~\ref{drift_het17}, \textit{red bars}).  Omitting heterogeneity as above reduced average performance to $77.38\%$ (Figure~\ref{drift_het17}, \textit{blue bars}).   

\begin{figure}
  \centering
  \includegraphics[width=2.0in]{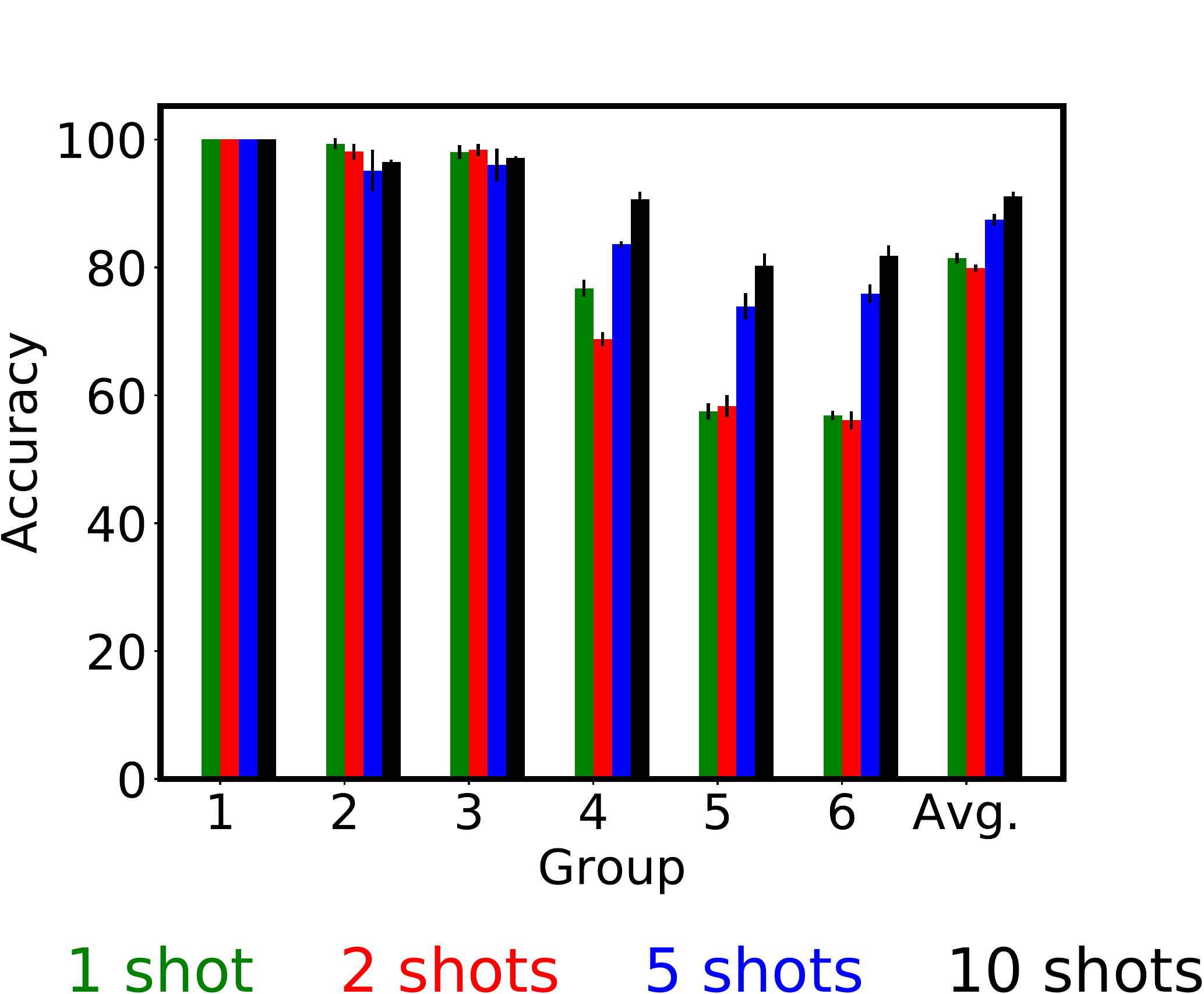}
  \caption{ Sample classification accuracies from Batch 7 of the UCSD chemosensor drift dataset using an online learning paradigm.  The six groups (odorants, with uncontrolled concentrations) were trained sequentially in the order depicted using 1, 2, 5 or 10 shots. After training each group, test samples from all six groups were presented, with samples from  yet-untrained groups being correctly classified as \textit{none of the above}. \textit{Avg.} depicts classification performance averaged across the six individual assessments.}
  \label{drift_all7}
\end{figure}

We then trained the network using two-shot, five-shot, and 10-shot online learning protocols. Training trials were grouped by odorant identity to demonstrate online learning (i.e., not intercalated); concentrations again were uncontrolled.  Classification accuracy improved substantially with the additional training (Table~\ref{tab:perf.}, Figure~\ref{drift_all7}) yielding a maximum of $91.10\%$ average accuracy for 10-shot training.  The 10-shot training set comprised $1.7\%$ of the Batch 7 data.  
 
\subsection{Forest type spectral maps}
Despite being inspired by the neural circuitry of the olfactory bulb, this network was expected to perform comparably well on datasets exhibiting structural properties similar to odorant stimuli:  relatively high dimensionalities without low-dimensional structure such as that exhibited by visual images. To demonstrate this, and to test the capacities of our preprocessors to appropriately regularize the statistical structures of non-chemosensory datasets, we tested the same network utilized above on two additional datasets.

We first tested the algorithm's performance on a 27-dimensional dataset of hyperspectral mapping data derived from ASTER satellite imagery, intended to identify four classes of Japanese forest cover \cite{johnson_using_2012, noauthor_uci_nodate-1}. The network was expanded from 16 input dimensions (for the UCSD dataset) to 27 input columns to match dataset dimensionality, and included $200$ granule cells per sensor. Despite substantial differences in signal statistics, our preprocessor cascade regularized the input distribution and achieved near-uniform interneuron recruitment (Figure~\ref{all_pre}b,e,h).

\begin{figure}
\centering
\subfloat[]{\includegraphics[width=0.45\linewidth]{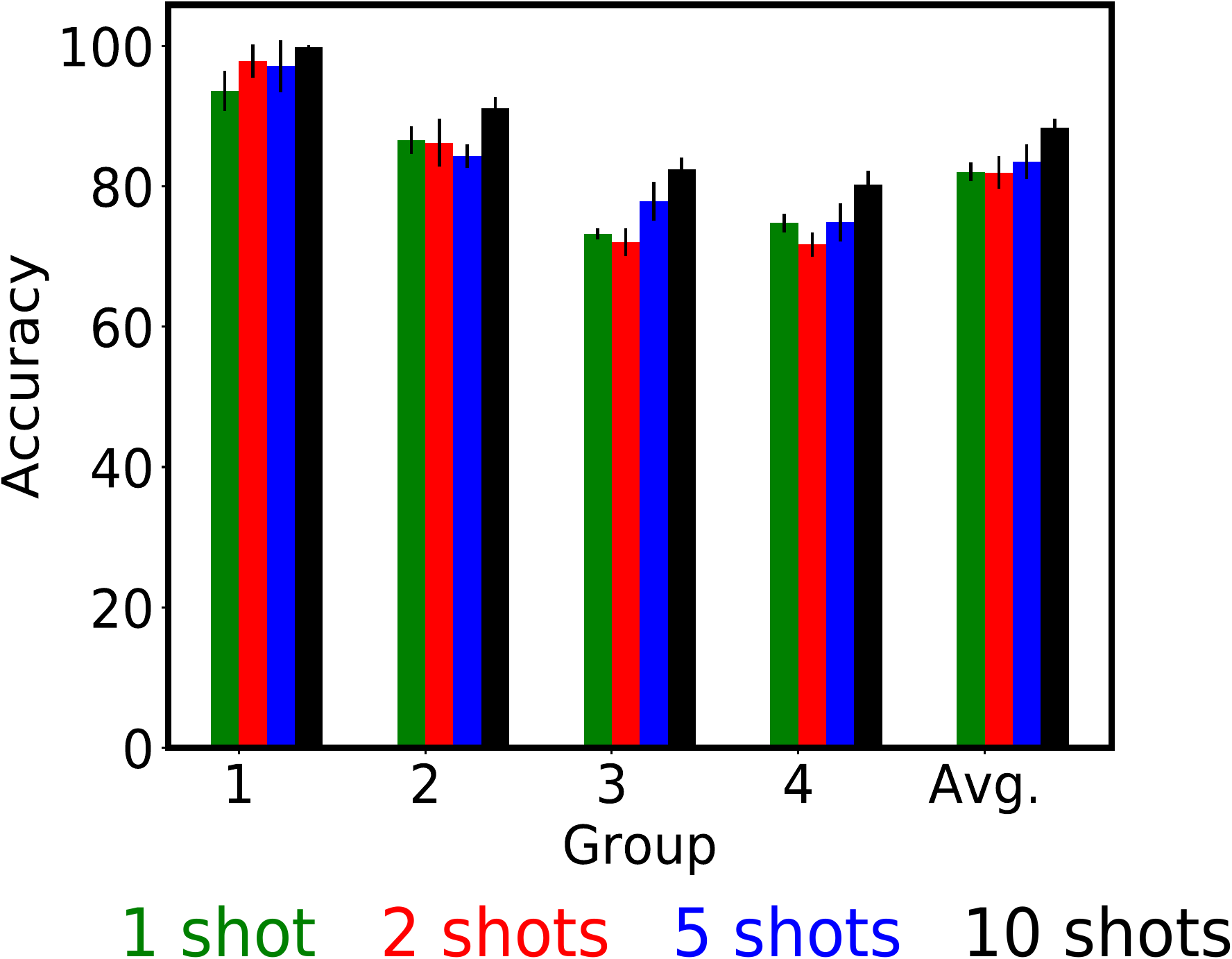}}
\hspace{0.25in}
\subfloat[]{\includegraphics[width=0.45\linewidth]{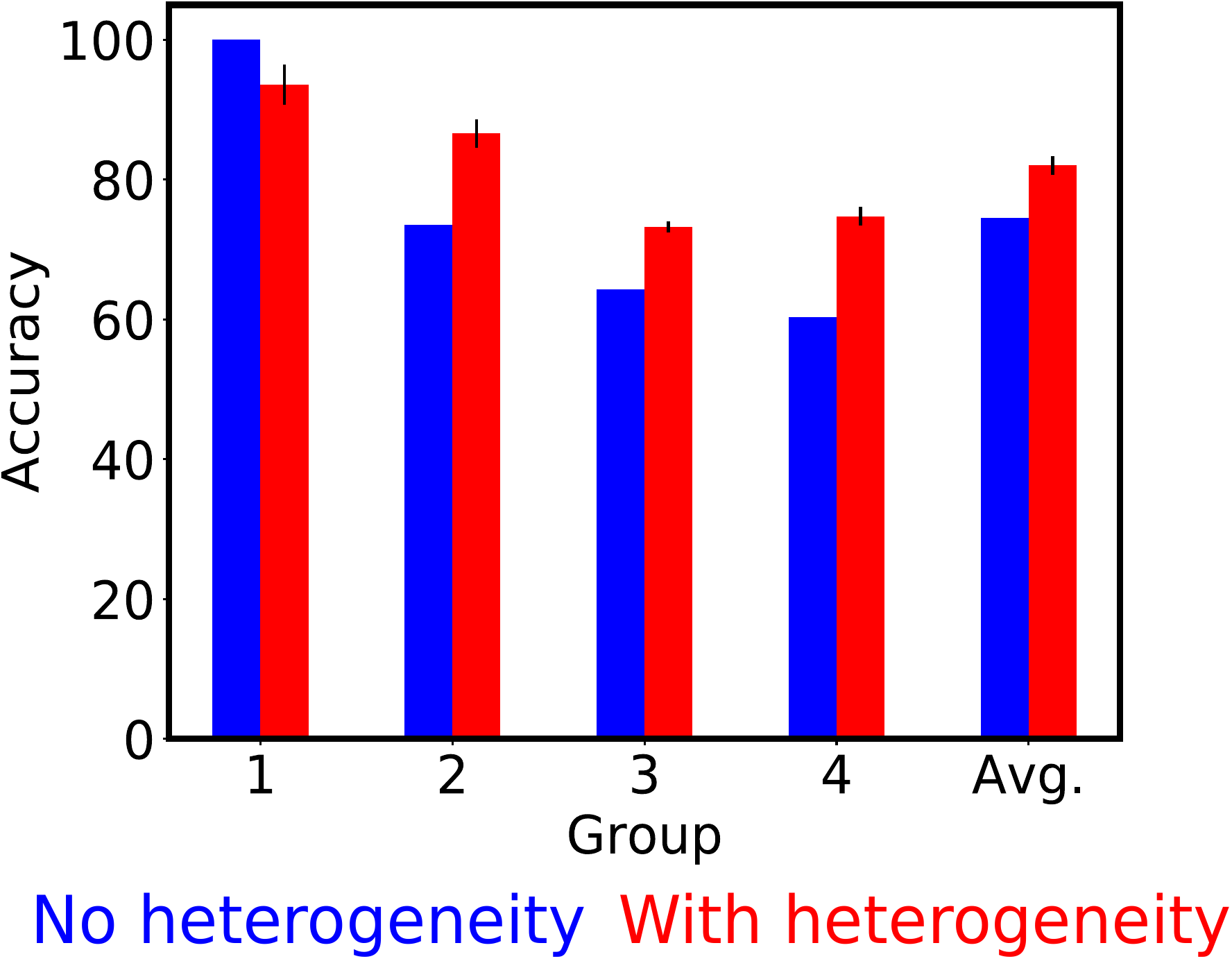}}
\caption{Sample classification accuracies based on the spectral attributes of four types of forest cover, using an online learning paradigm. (a) Performance comparison of 1, 2, 5, and 10-shot learning. (b) Performance comparison of one-shot learning with or without the inclusion of network parameter heterogeneity.  \textit{Avg.} depicts classification performance averaged across the four individual assessments.}
\label{forest}
\end{figure}

We trained the network with one shot of each of the four forest types; the training set consequently comprised $0.76\%$ of the data (4 of 523 samples), and the test set comprised $89.24\%$ (463 of 523 samples). The average classification accuracy across all groups was $82.03\%$ (Table~\ref{tab:perf.}, Figure~\ref{forest}a).  Because of the special status of the \textit{Other} group, \textit{Other} classifications were pooled with \textit{none of the above} classifications after the network was trained on all four groups.  Performance improved after two-, five-, and 10-shot training, reaching $88.39\%$ after ten-shot learning (the training set here comprised $7.65\%$ of the data). When we omitted network heterogeneities, as with the UCSD chemosensory dataset, the average accuracy for one-shot learning dropped from $82.03\%$ to $74.53\%$ (Figure~\ref{forest}b).  

\subsection{Species-specific anuran calls}
Finally, we also tested the algorithm on an implicitly hierarchical classification task using a dataset derived from a corpus of recordings of vocalizations from ten anuran species. As detailed above, the dataset comprised 22 mel frequency cepstral coefficients describing the acoustic features of these call syllables.  We sought to identify the animal species, but also the genus and family, associated with each call. To do this, we deployed a network with hyperparameters identical to those used in prior datasets, with two exceptions.  First, the network was necessarily sized for the 22 input dimensions of the dataset.  Second, the number of interneurons was expanded to 300 per sensor; this was necessary in order to adequately learn all ten classes without the adaptive network expansion function of the fully intact network \cite{imam_rapid_2019}.  As with the earlier datasets, preprocessing yielded a consistent statistical distribution of input amplitudes and a near-uniform recruitment of interneurons (Figure~\ref{all_pre}c,f,i; Table~\ref{tab:perf.}).  

\begin{figure}
\centering
\subfloat[Species]{\includegraphics[width=0.45\linewidth]{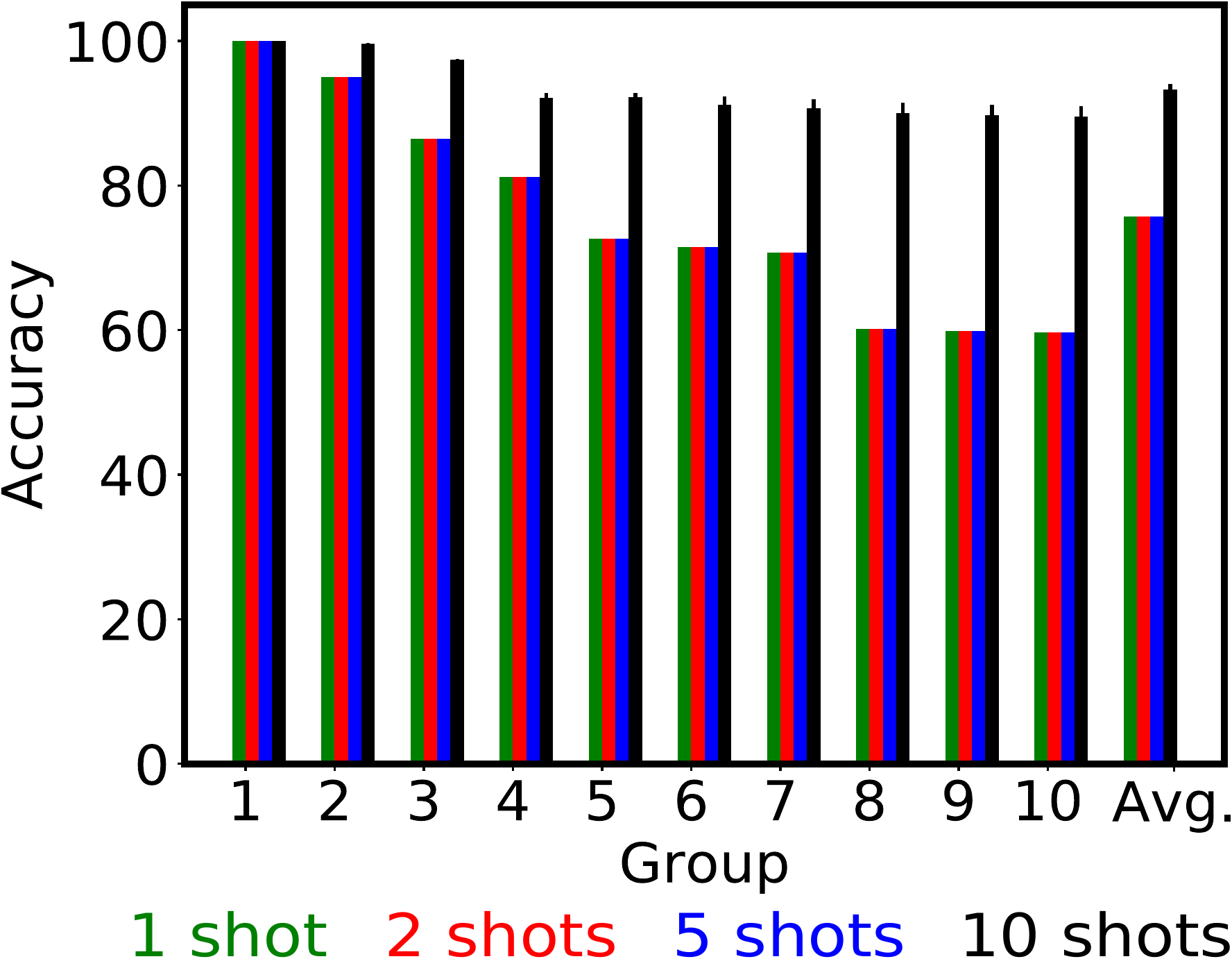}}
\hspace{0.25in}
\subfloat[Species]{\includegraphics[width=0.45\linewidth]{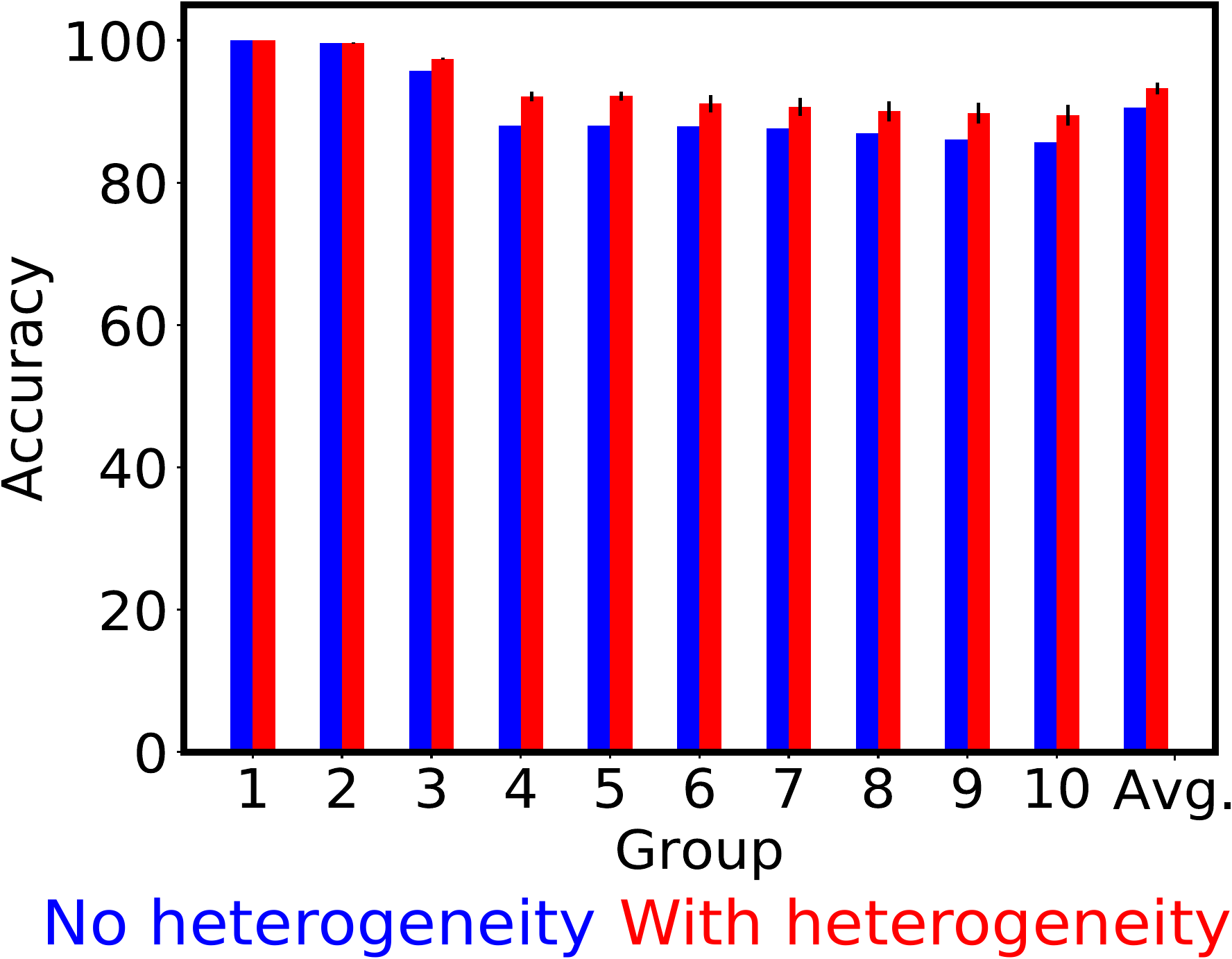}}

\subfloat[Genus]{\includegraphics[width=0.45\linewidth]{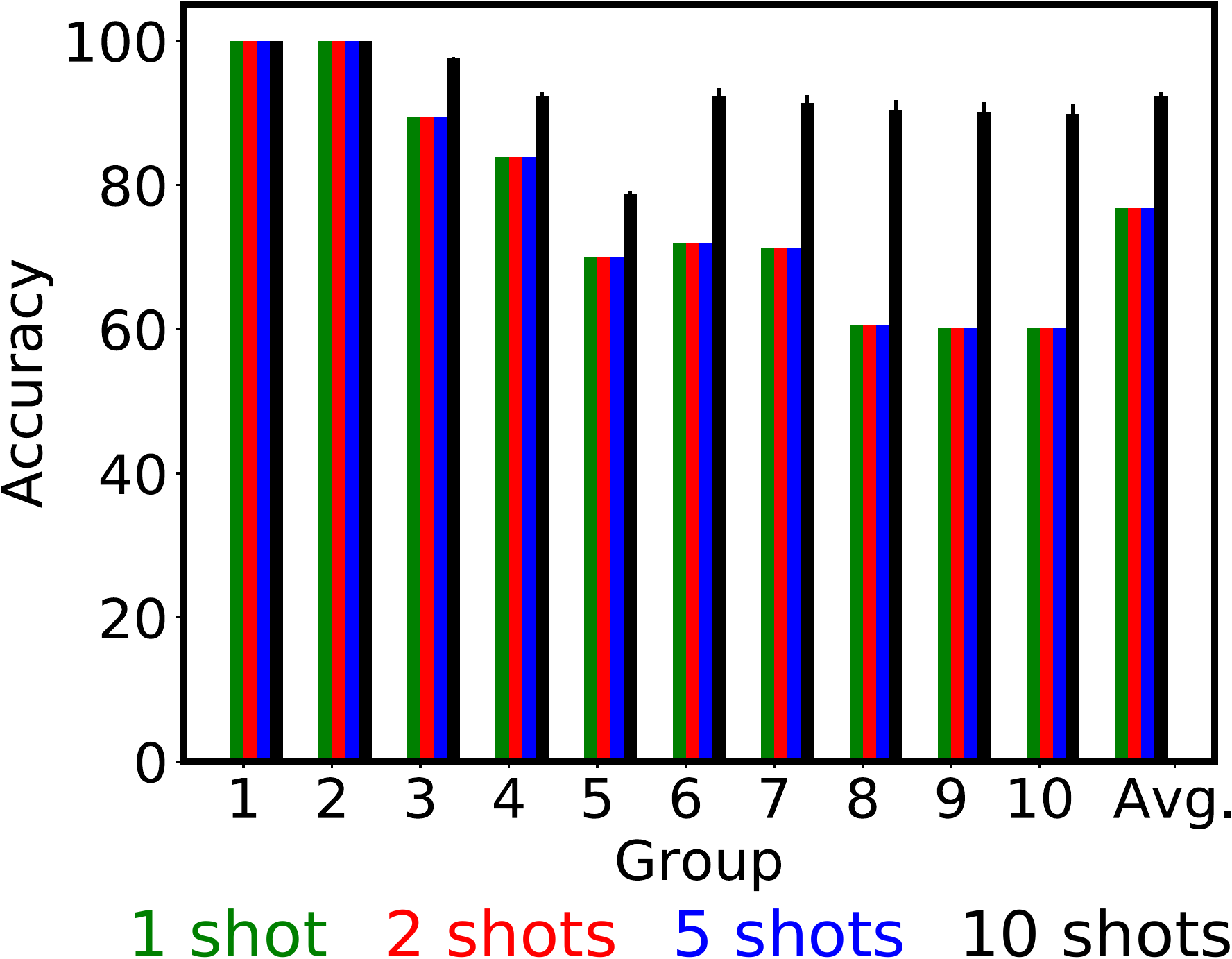}}
\hspace{0.25in}
\subfloat[Genus]{\includegraphics[width=0.45\linewidth]{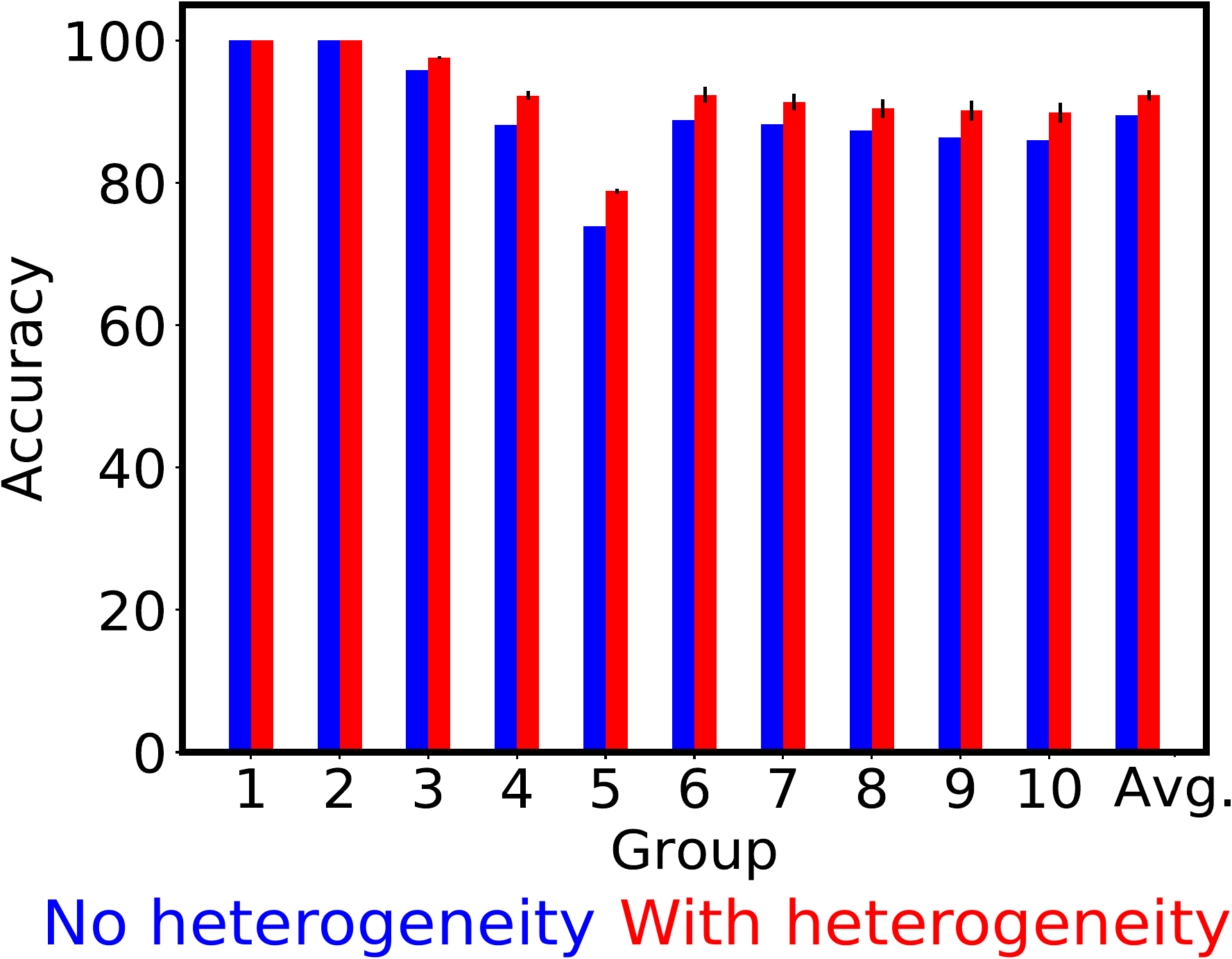}}

\subfloat[Family]{\includegraphics[width=0.45\linewidth]{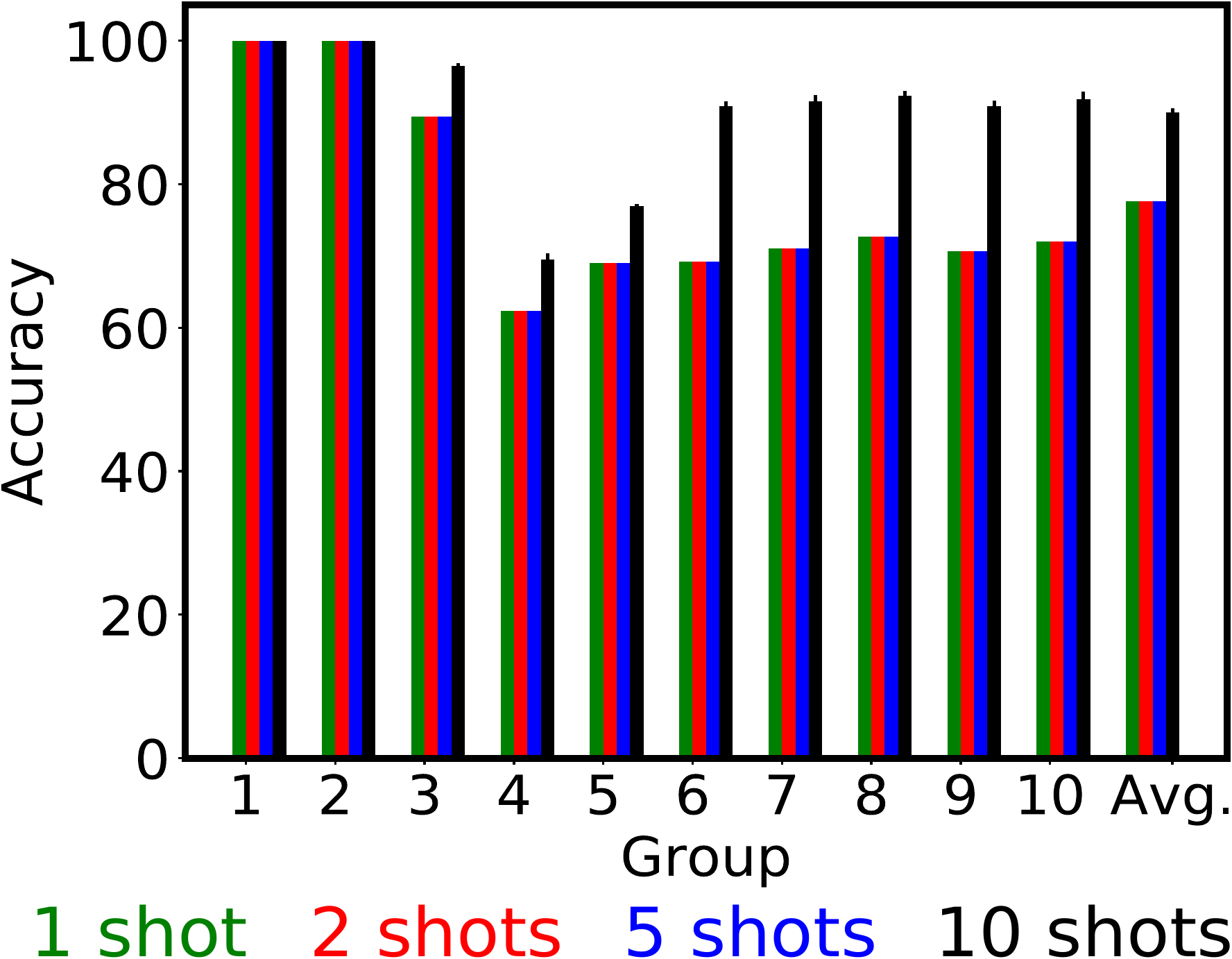}}
\hspace{0.25in}
\subfloat[Family]{\includegraphics[width=0.45\linewidth]{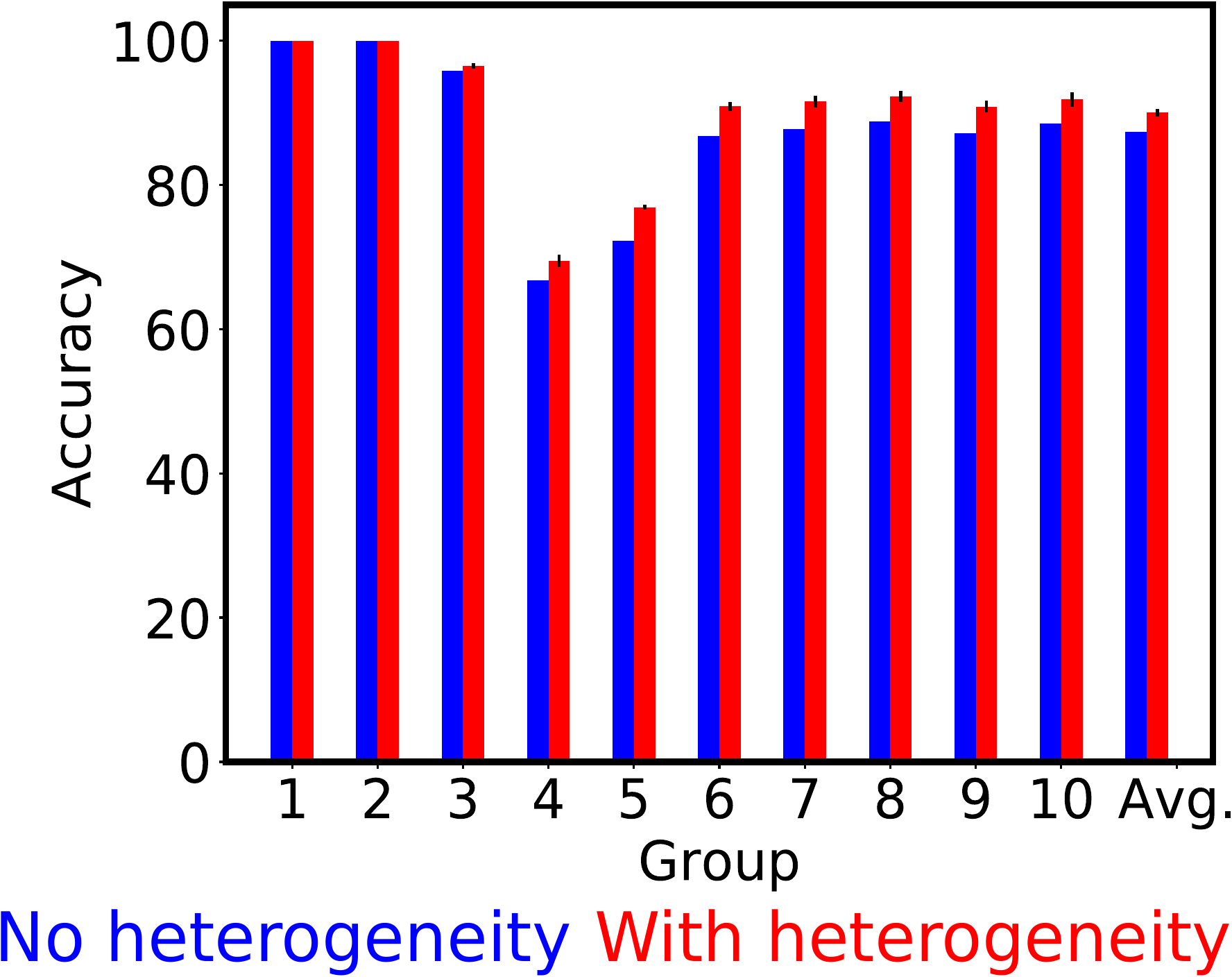}}

\caption{Classification performance using the anuran species call dataset. (a,c,e) Network performance using 1, 2, 5, or 10-shot learning.  (b,d,f) Comparison of 10-shot learning performance without and with parameter heterogeneity.  (a,b) Calls classified according to species.  (c,d) Calls classified according to genus.  (e,f) Calls classified according to family.  \textit{Avg.} depicts classification performance averaged across the ten individual assessments (\textit{groups}).}

\label{anuran}
\end{figure}

One-shot online learning of the ten groups (species) in this dataset yielded somewhat poorer classification accuracy than in the previous datasets tested; the accuracy across groups averaged $75.72\%$ (Figure~\ref{anuran}a, Table~\ref{tab:perf.}), with the training set size comprising just $0.14\%$ of the dataset (10 of 7195 samples). Expanding to two- and five-shot training produced little improvement.  However, expansion to 10-shot training improved classification accuracy to $93.25\%$, with the training set comprising $1.39\%$ of the data (100 of 7195 samples). Removing parameter heterogeneity reduced 10-shot classification performance to $90.54\%$ (Figure ~\ref{anuran}b).  

Finally, we assessed classification performance with respect to the eight anuran genera and four families embedding the ten species on which the network was trained.  No additional training or network design was performed; output was simply reclassified with respect to these higher cladistic levels. Performance on this classification task largely tracked that of classifying by species (Table~\ref{tab:perf.}), with accuracy increasing substantially given 10-shot training (Figure ~\ref{anuran}c,e) and being modestly impaired by the removal of network heterogeneity (Figure ~\ref{anuran}d,f).  This implicit capacity to respect hierarchical similarity relationships is a substantial benefit of the generalized, similarity-representing variant of this algorithm as described herein.  

\begin{table}[!htb]
  \caption{Average classification performance on each experimental dataset, using an online learning paradigm designed for online learning and assessed using an intermediate classification metric (see text for details).}
  \label{tab:perf.}
  \begin{tabular}{ccccc}
    \toprule
     & 1 shot & 2 shots & 5 shots & 10 shots\\
    \midrule
    Batch 1 & $\underset{\pm 1.32}{\textbf{96.00}}$ &  &  & \\
    \addlinespace[0.2cm]
    Training set size & $1.35\%$ &  &  & \\
    \midrule
    Batch 7 & $\underset{\pm 0.83}{81.42}$ & $\underset{\pm 0.54}{79.94}$ & $\underset{\pm 0.91}{87.43}$ & $\underset{\pm 0.75}{\textbf{91.10}}$\\
    \addlinespace[0.2cm]
    Training set size & $0.17\%$ & $0.33\%$ & $0.83\%$ & $1.7\%$ \\
    \midrule
    Forest & $\underset{\pm 1.33}{82.03}$ & $\underset{\pm 2.33}{81.96}$ & $\underset{\pm 2.47}{83.53}$ & $\underset{\pm 1.29}{\textbf{88.39}}$\\
    \addlinespace[0.2cm]
    Training set size & $0.76\%$ & $1.53\%$ & $3.82\%$ & $7.65\%$ \\
    \midrule
    Anuran (species) & $\underset{\pm 0.}{75.72}$ &  $\underset{\pm 0.}{75.72}$& $\underset{\pm 0.}{75.72}$& $\underset{\pm 0.81}{\textbf{93.25}}$\\
    Anuran (genus) & $\underset{\pm 0.}{76.74}$ & $\underset{\pm 0.}{76.74}$ & $\underset{\pm 0.}{76.74}$ & $\underset{\pm 0.71}{\textbf{92.26}}$\\
    Anuran (family) & $\underset{\pm 0.}{77.63}$ & $\underset{\pm 0.}{77.63}$ & $\underset{\pm 0.}{77.63}$ & $\underset{\pm 0.56}{\textbf{90.04}}$\\
    \addlinespace[0.2cm]
    Training set size & $0.14\%$ & $0.28\%$ & $0.69\%$ & $1.39\%$ \\
  \bottomrule
\end{tabular}
\end{table}

\section{Discussion}
Learning in the wild comprises an aspirational set of capacities for artificial networks that reflect the performance of biological systems operating in natural environments. Most of the difficult challenges arise from a sharply limited ability to regulate the stimuli presented by the external environment, whether in their unpredictable diversity, their interference with one another, or their intrinsic variances.  The specific capacities that we have required of our generalized algorithm include the following:  
\begin{itemize}
   \item It must be robust to \textit{wild}, poorly-matched inputs without resorting to hyperparameter re-tuning.
   \item It must be robust to environmental and stimulus variance, including unpredictable stimulus intensities (e.g., odorant concentrations), other forms of stimulus heterogeneity, and the effects of environmental temperature and humidity.
   \item It must exhibit concentration tolerance where appropriate, and also provide an estimate of concentration.
   \item It must be robust to missing or noisy sensor data, and to unlabelled training sets.
   \item It must exhibit rapid, semi-supervised or unsupervised, one- or few-shot learning of novel stimuli.
   \item It must support online learning (no catastrophic forgetting, no need to store trained data).
   \item It must adapt to sensor drift owing to time and/or contamination \cite{borthakur_spike_2019}.
   \item It must provide a \textit{none of the above} option during classification (classifier confidence) \cite{borthakur_spike_2019}.
   \item It must be able to identify the signatures of known inputs despite substantial interference from background stimuli (whether previously or simultaneously delivered).
\end{itemize}
The initial implementation of this algorithm \cite{imam_rapid_2019} exhibited a majority of these properties. In subsequent implementations, we have harnessed the network's rapid learning capabilities to achieve a practical solution to the problem of sensor drift \cite{borthakur_spike_2019} and generalized the algorithm to embed an explicit representation of similarity so as to enable support for hierarchical clustering. A preliminary example of this capacity is illustrated here in the classification of anuran calls with respect to species, genus, and family.  This generalized implementation of the algorithm, however, becomes necessarily more sensitive to the statistical structure of sensory inputs.  We here have outlined a signal conditioning solution in which \textit{wild} sensory inputs are regularized by a series of preprocessors modeled on the features and circuits of the olfactory bulb glomerular layer.  Consequently, a single instantiated network is capable of productively learning and classifying widely heterogeneous sets of input stimuli.  

Data normalization in some form is a common procedure in non-spiking neural networks \cite{ioffe_batch_2015, bjorck_understanding_2018}. We here sought to implement a data regularization procedure for spiking neural networks that was compatible with rapid learning, localized brain-mimetic computational principles, and \textit{learning in the wild} constraints. Notably, under these constraints, samples may be rare, and batch sizes small, such that aggregate data features such as means and standard deviations are difficult to ascertain.  We further sought to ensure that single instantiated networks could effectively learn and classify a wide diversity of datasets.  The successive preprocessors described herein transformed four different datasets with different patterns of internal sample diversity into a common statistical form, such that the same network could effectively operate on them all without the need for hyperparameter retuning.  

The final preprocessor in the sequence, heterogeneous duplication (Figure ~\ref{het_dup}), is a statistical regularization algorithm based on the properties of sparse random projections.  Interestingly, its implementation closely adheres to an anatomical circuit motif within olfactory bulb intraglomerular networks \cite{gire_etcells_2012}, to which function has yet to be attributed.  The need for statistical regularization of input patterns in this way has not yet been recognized in the literature on biological olfaction (except in the specific case of concentration), so it is an interesting possibility that this network motif may present a solution to a previously unrecognized neurophysiological problem.  

The simulations in this paper concern the initial preprocessing steps and first feed-forward projection of the biomimetic algorithm (Figure~\ref{schematic}; corresponding to the \textit{EPLff} component described in \cite{borthakur_spike_2019}), omitting the dynamical spike timing-based attractor functionality of the full network \cite{imam_rapid_2019} in favor of a closer examination of preprocessor properties.  Accordingly, the metrics of greatest interest are the uniformity of interneuron recruitment and a preliminary estimate of classification performance based on the Hamming distances calculated between interneuronal activation patterns.  The latter metric, in particular, should not be confused with the performance of a fully implemented brain-mimetic  implementation \cite{imam_rapid_2019}; obtaining optimized classification accuracy was not the primary purpose of this reduced network.  Among other limitations, the Hamming distance metric cannot accommodate the adaptive network expansion (ANE) method, by which new interneurons are dynamically added to the network after the fashion of adult neurogenesis in the olfactory bulb \cite{imam_rapid_2019}, because ANE alters the dimensionality of the space in which the Hamming distance is calculated. Owing to the absence of ANE, the present network's performance begins to drop off as the number of learned stimuli increases; this can be observed in Figures~\ref{drift_het17}--\ref{anuran}.  Reported average performance values, accordingly, are underestimates of intact network performance.

The central message of the present work is that a series of preprocessing steps, modeled after particular attributes of the mammalian olfactory bulb, successfully conditions statistically diverse input signals from both chemosensory and nonchemosensory sources, such that a single instantiated, parameterized network can rapidly learn and successfully classify these signals.  We have termed this robustness to uncontrolled environmental variance \textit{learning in the wild}.  This is a critical capability for field-deployed devices expected to process and identify similarly diverse sensory signatures within unregulated environments.  Moreover, as with the intact network \cite{imam_rapid_2019}, these preprocessor algorithms were implemented using localized computational and plasticity rules and hence are amenable to implementation on neuromorphic hardware platforms \cite{davies_loihi:_2018}.  

%
\begin{acks}
This work was supported by a Cornell University Sage fellowship to AB and NIH/NIDCD awards DC014367 and DC014701 and an Intel Neuromorphic Research Community faculty award to TAC. 
\end{acks}

%
\bibliographystyle{ACM-Reference-Format}
\bibliography{sample-base}

\end{document}